\documentclass{article}
\usepackage{arxiv}
\usepackage[utf8]{inputenc} 
\usepackage[T1]{fontenc}    
\usepackage{hyperref}       
\usepackage{url}            
\usepackage{booktabs}       
\usepackage{amsfonts}       
\usepackage{nicefrac}       
\usepackage{microtype}      
\usepackage{lipsum}
\usepackage{hyperref}
\usepackage{amsmath,amsfonts,amsthm,amssymb,bm,bbm} 

\newcommand{\R}{\mathbb{R}}
\makeatother
\usepackage{dirtytalk}
\usepackage{float}
\usepackage{siunitx}
\usepackage{booktabs}
\usepackage{multirow}
\usepackage{tabulary}
\usepackage{graphicx}

\title{AutoDiscern: Rating the Quality of Online Health Information with Hierarchical Encoder Attention-based Neural Networks}

\author{
  Laura Kinkead\thanks{Biomedical Informatics, University Hospital of Zurich, Zurich, Switzerland} \\
  Department of Quantitative Biomedicine\\
  University of Zurich\\
  Schmelzbergstrasse 26, Zurich, CH \\
  \texttt{laura.kinkead@uzh.ch} \\
   \And
  Ahmed Allam\thanks{Biomedical Informatics, University Hospital of Zurich, Zurich, Switzerland} \\
  Department of Quantitative Biomedicine\\
  University of Zurich\\
  Schmelzbergstrasse 26, Zurich, CH \\
  \texttt{ahmed.allam@uzh.ch} \\
   \And
  Michael Krauthammer\thanks{Biomedical Informatics, University Hospital of Zurich, Zurich, Switzerland} \thanks{Department of Pathology, Yale University School of Medicine, New Haven, USA} \\
  Department of Quantitative Biomedicine\\
  University of Zurich\\
  Schmelzbergstrasse 26, Zurich, CH \\
  \texttt{michael.krauthammer@uzh.ch} \\
}

\begin{document}
\maketitle

\begin{abstract} 
\textbf{Background} 
Patients increasingly turn to search engines and online content before, or in place of, talking with a health professional. Low quality health information, which is common on the internet, presents risks to the patient in the form of misinformation and a possibly poorer relationship with their physician. To address this, the DISCERN criteria (developed at University of Oxford) are used to evaluate the quality of online health information. However, patients are unlikely to take the time to apply these criteria to the health websites they visit. 

\textbf{Methods} 
We built an automated implementation of the DISCERN instrument (Brief version) using machine learning models. We compared the performance of a traditional model (Random Forest) with that of a hierarchical encoder attention-based neural network (HEA) model using two language embeddings, BERT and BioBERT.

\textbf{Results} 
The HEA BERT and BioBERT models achieved average F1-macro scores across all criteria of 0.75 and 0.74, respectively, outperforming the Random Forest model (average F1-macro = 0.69). 
Overall, the neural network based models achieved 81\% and 86\% average accuracy at 100\% and 80\% coverage, respectively, compared to 94\% manual rating accuracy. The attention mechanism implemented in the HEA architectures not only provided 'model explainability' by identifying reasonable supporting sentences for the documents fulfilling the Brief DISCERN criteria, but also boosted F1 performance by 0.05 compared to the same architecture without an attention mechanism. 

\textbf{Conclusions} 
Our research suggests that it is feasible to automate online health information quality assessment, which is an important step towards empowering patients to become informed partners in the healthcare process. 

\end{abstract}


\section*{Introduction}
\subsection*{Background}
Patients often turn to search engines and online content before, or in place of, talking with a health professional \cite{Hesse2005}. 
However, online health information is not regulated, and prior studies have found wide variations in information quality \cite{Fahy2014}. Poor risk communication, biased writing, and lack of transparency about the source of the information plague online health texts \cite{Zhang2015,Saunders2018}. This presents a real risk to patients, in the form of misinformation \cite{Murray2003,Allam2014, Ludolph2016} and negatively affecting their interactions with health care providers \cite{Iverson2008, Wald2007}.

In response to this problem, many organizations, such as the Health on the Net Organization, the Journal of the American Medical Association, and the National Health Service of the UK, have established guidelines for assessing the quality of online health information \cite{Risk2001}. These guidelines describe a set of criteria an article must meet to be considered of high quality. It is worth noting that \emph{quality} is distinct from \emph{accuracy}. While these guidelines check for indicators of well written, unbiased, and evidence based articles, they do not attempt to verify the scientific accuracy of the information (a significantly more challenging problem). Similarly, the concept of quality is also distinct from that of \emph{credibility}, or how likely a reader is to believe the information. The propensity to which readers believe the content they consume is influenced not only by information accuracy, but also structural aspects of the media, such as a website's design, appearance, and overall readability \cite{Viviani2017}. Thus, quality guidelines form a basis by which systems may affect individual's perceptions of credibility, without breaching into the field of information accuracy assessment.

The implementation strategies of these quality guidelines so far fall into two categories: Distributed Guidelines and Centralized Approvers. 
However, both of these strategies have scalability issues that limit their reach and prevent them from broadly affecting patient information consumption \cite{Risk2001}. In the following section, we describe both of these implementation approaches in use today, and then describe a third solution that addresses the issue of scalabilty.

\paragraph*{Distributed Guidelines} One approach to helping patients find high quality health information is to develop a criteria and publish it as a public tool citizens can use. An example of this approach is the DISCERN instrument \cite{Charnock1999}. The DISCERN instrument's criteria are specifically designed to be able to be understood and applied by any lay person; no medical knowledge is required.
This implementation approach puts a significant burden on the patient. For this approach to be successful, the patient has to be aware of the guideline, learn how to evaluate the criteria, and take considerable time to apply the guidelines to every website the patient encounters. 

\paragraph*{Centralized Approvers} The second implementation approach in use today is Centralized Approvers. In this approach, an organization manually assesses web pages for health information quality. An example of this approach is the Health on the Net Foundation, which developed the HONcode guidelines. It assesses websites for quality, and allows those that pass their criteria to display a HONcode badge on their webpage \cite{Boyer2014}. 
A variant on this approach is to register all manually approved content in a centralized repository. Patients can search the repository with the confidence that all listed sites have been vetted for quality. 

The Centralized Approver approach is not scalable in the face of a massive and rapidly growing internet. Quality assessment is a costly manual process. Not only do new pages need to be evaluated, but previously-evaluated pages need to be re-evaluated on a regular basis in case of content changes \cite{Risk2001}. 


\paragraph*{Automated Assessment} 
An automated quality assessment process is key to providing the public with scalable tools for assessing online health information quality.

Initial attempts to automate the assessment of health information used simplistic approaches, such as readability scores, and did not capture more complex issues with health information, such as tone and bias \cite{Saunders2018}. A machine learning model developed by the HON organization showed promising but limited initial results \cite{Boyer2015}. But with the recent developments in machine learning and natural language processing methods, there is a renewed opportunity for tackling this problem. Neural Language Models have been successfully applied in many domains, including translation, question answering, and many more \cite{Vaswani2017, Luong2015, Wolf2019, Ruder2019}, capturing details and nuances in language that made information quality assessment an expensive manual process for so long.

\subsection*{Research Objectives}
In this research, we study and develop machine learning models to automate the application of the DISCERN instrument. The DISCERN instrument was developed by Charnock et al. \cite{Charnock1999} at Oxford University and funded by the National Health Service (UK). The instrument consists of 15 questions to help a lay-person to evaluate the quality of online health information regarding treatment options. The validity of the DISCERN instrument has been evaluated in multiple studies, and is commonly used among researchers \cite{Rees2002}. The DISCERN instrument suffers from the same sustainability issues as all distributed guidelines do: patients are unlikely to take the time to apply this criteria to each website they find. In this study, we built and evaluated machine learning models for the automated annotation of the Brief DISCERN criteria \cite{Khazaal2009}. 
We focus on the Brief DISCERN criteria \cite{Khazaal2009}, which is a 6 question subset of the DISCERN crieria that has been shown to capture the quality of health information as reliably as the complete DISCERN instrument. Separate models were developed and tested for each of the five Brief DISCERN questions (one question, Q13, was excluded due to low interrater reliability).
We compared the use of traditional machine learning (Random Forest) with feature engineering vs. hierarchical encoder attention-based neural network (HEA) models. We also compared the performance of neural models with the attention mechanism (HEA) and without it (HE). Additionally, for both neural architectures, we experimented with the use of two pre-trained neural language models BERT \cite{devlin2018bert} and BioBERT \cite{10.1093/bioinformatics/btz682} as embeddings in the HEA and HE models. Thus, in total, we trained and compared 5 different architectures: RF, HE+BERT, HE+BioBERT, HEA+BERT, and HEA+BioBERT. 

\section*{Methods}
\subsection*{Data Collection}
Using Google Trends, we identified breast cancer, arthritis, and depression as medical topics with the highest search volume since 2004. Using Google and Yahoo search engines, we identified a total of 269 Web pages (HTML articles) with a focus on treatment choices and options across the 3 topics. Two raters (master’s students) were trained for 2 months on using the DISCERN instrument and scoring platform. Both raters scored all articles on DISCERN's 5 point scale. Interrater agreement for the DISCERN criteria was adequate to high, ranging between 0.61–0.91 as measured by the Krippendorf score. The process of building the training corpus is described in more detail in \cite{Allam}. The dataset is described in Table \ref{table:dataset_stats}. 

\begin{table}[h!]
\caption{Description of the dataset by health topic.}
\label{table:dataset_stats}
\begin{center}
    \normalsize
      \begin{tabular}{lrrr}
      \toprule
        Topic                & Breast Cancer  & Arthritis & Depression \\ 
        \midrule
        Number of Articles        &    79     &    88     &   102    \\
        Number of Sentences       & 10,170    & 10,950    &   13,790 \\
        Number of Tokens          & 125,891   & 129,759   &  160,597 \\
        Avg Sentences per Article &   129     &   124     &   135    \\
        Avg Tokens per Article    & 1,549     & 1,475     & 1,574    \\
        & & & \\
        Positive Class Prevelance & & & \\
        \midrule
        Q4: References            &   13\%    &   14\%  &  14\%    \\
        Q5: Date                  &   20\%    &   26\%  &  24\%    \\
        Q9: How Treatment Works   &   85\%    &   28\%  &  52\%    \\
        Q10: Treatment Benefits   &  89\%     &   80\%  &  65\%    \\
        Q11: Treatment Risks      &  63\%     &   16\%  &  33\%    \\
        \bottomrule
      \end{tabular}
    \end{center}
\end{table}

\subsection*{Data Preprocessing}
We converted the scores for each question in the DISCERN instrument, which ranges from 1-5, into a binary classification, where score 3-5 is passing and score 1-2 is failing the criteria. The texts from the HTML articles were extracted and cleaned using the beautifulsoup library \cite{richardson2007beautiful}.

\subsection*{Modeling Approach}
We converted the scores for each question in the DISCERN instrument, which ranges from 1-5, into a binary classification, where score 3-5 is passing and score 1-2 is failing the criteria. The texts from the HTML articles were extracted and cleaned using the beautifulsoup library \cite{richardson2007beautiful}.
In this work we focus on the Brief DISCERN criteria \cite{Khazaal2009}, which is a 6 question subset of the DISCERN crieria that has been shown to capture the quality of health information as reliably as the complete DISCERN instrument. Separate models were developed and tested for each of the 5 Brief DISCERN questions (one question, Q13, was excluded due to low interrater reliability).

\subsection*{Neural Network Model}

\begin{figure}[h!]
    \caption{Overview of the HEA neural network architecture. In lieu of traditional feature engineering, the HEA architecture learns representations at the word, sentence, and document level before making a classification. Word representations are generated by the pre-trained BERT embedder. An attention mechanism aids in learning a document representation from amongst many sentences.}
  \centering
      \includegraphics[width=\textwidth]{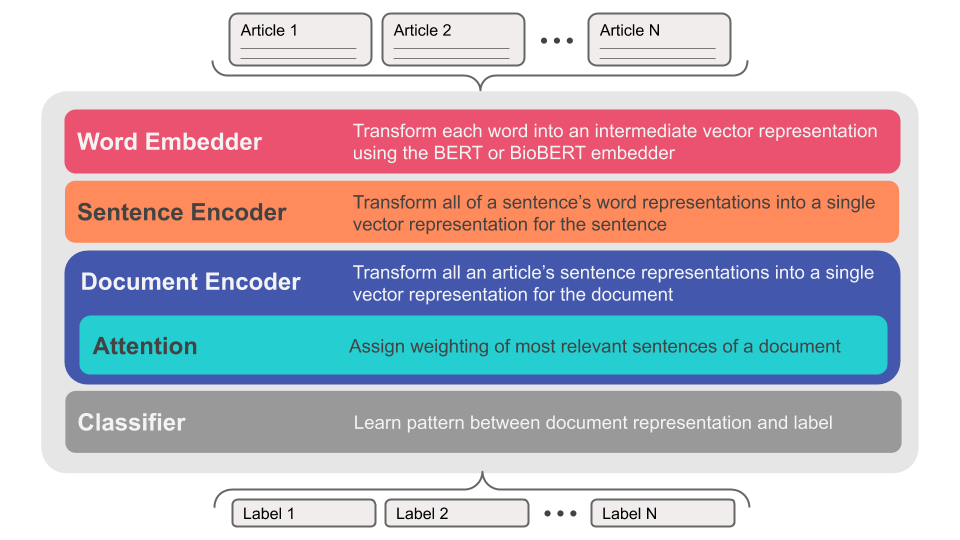}
  \label{fig:architecture_overview}
\end{figure}

We designed and implemented a Hierarchical Encoder Attention-based (HEA) model in PyTorch \cite{Paszke2017AutomaticDI} taking into consideration the structure of our problem and the limits of our training data. 
The model architecture design is primarily motivated by the intrinsic hierarchy of the documents (i.e. sequences of word/tokens represent a sentence, and sequences of sentences represent a document). In addition, our  attention-based  modeling architecture reflects the property that passing or failing the DISCERN criteria depends on only small fragments throughout the article. This architecture enables the model to "pay attention" to single sentences within a larger article. 

\begin{figure}
  \centering
      \includegraphics[width=\textwidth]{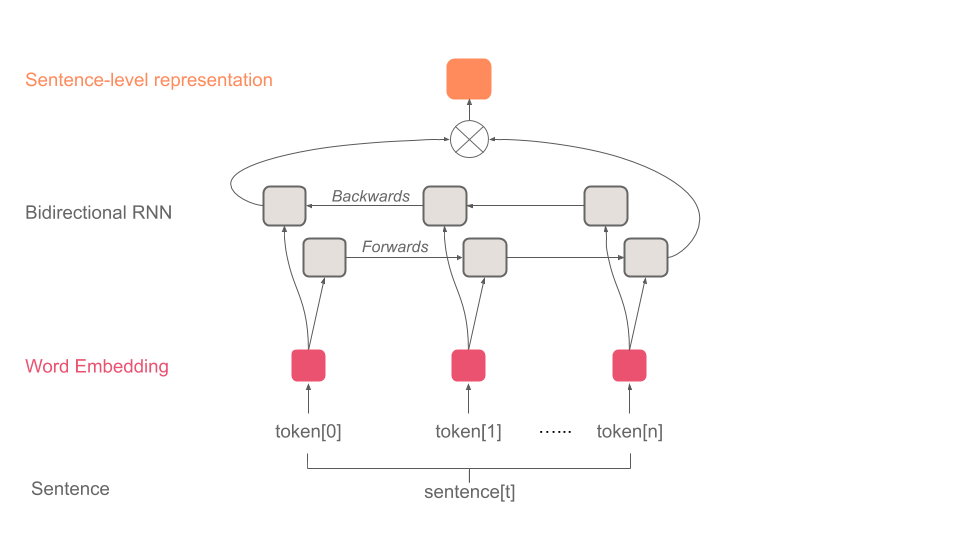}
  \caption{HEA's \emph{SentEncoder} architecture for computing sentence embedding.}
  \label{fig:architecture_sent}
\end{figure}

\begin{figure}
  \centering
      \includegraphics[width=\textwidth]{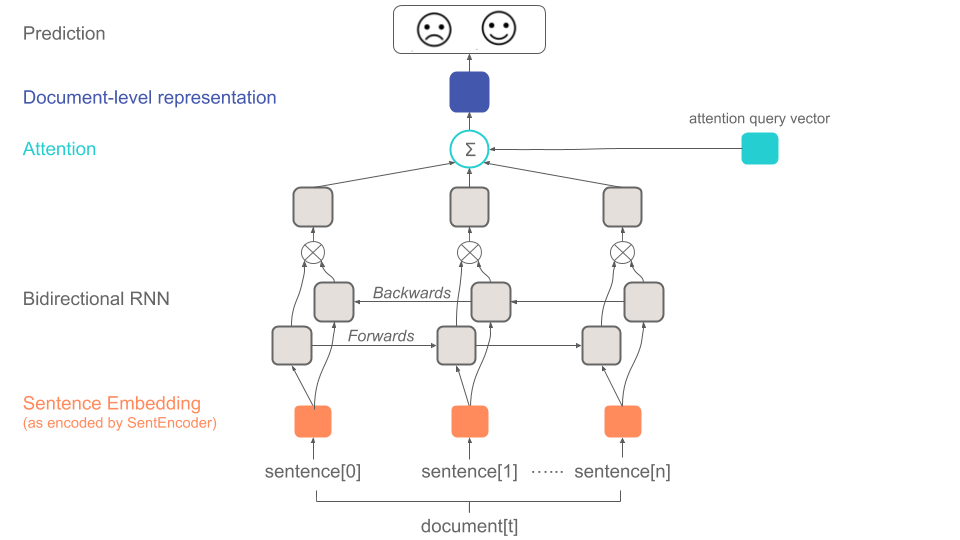}
  \caption{Model architecture for converting a document's sentence embeddings into a document prediction.}
  \label{fig:architecture_doc}
\end{figure}

HEA's architecture is composed of a hierarchical structure with two encoders and a classifier (Figure \ref{fig:architecture_overview}). The first encoder is a sentence encoder \emph{SentEncoder} (Figure \ref{fig:architecture_sent}) which is based on a bidirectional recurrent neural network (RNN) that encodes each sentence (i.e. sequence of tokens) into a dense vector representation. The second encoder is a document encoder \emph{DocEncoder} (Figure \ref{fig:architecture_doc}) which is also based on a bidirectional RNN that encodes the sequence of sentences' representation (i.e. vectors computed from the first encoder) and uses an attention mechanism \cite{Vaswani2017} along with a global context vector to compute a dense vector representation for the whole document. A decoder/classifier maps the document's learned vector representation to the labels using an affine map followed by \emph{softmax} layer computing a probability distribution on the labels for the processed document. An overview of the HEA model architecture can be found in Figure \ref{fig:architecture_overview}.

\subsubsection*{Sentence Encoder (SentEncoder)}
Formally, given an input sentence $\underline{S} = [\overline{w}_1, \overline{w}_2, \cdots, \overline{w}_{T_S}]$ where $\overline{w}_t$ represents the token representation at position $t$ (i.e. 1-of-\emph{K} encoding where $K$ is the size of vocabulary $V$ -- the set of all tokens in the training corpus), a vanilla RNN will compute a hidden vector at each position (i.e. state vector $\overline{h}_{t}$ at position $t$), representing a history or context summary of the sequence using the input and hidden states vector form the previous steps. Equation \ref{eq:rnn_ht} shows the computation of the hidden vector $\overline{h}_t$ using the input $\overline{w}_t$ and the previous hidden vector $\overline{h}_{t-1}$ where $\phi$ is a non-linear transformation such as $ReLU(z)=max(0,z)$ or $tanh(z)=\frac{e^z - e^{-z}}{e^z + e^{-z}}$.
\begin{equation}
\overline{h}_{t}=\phi(\mathbf{W}_{hw}\overline{w}_{t}+\mathbf{W}_{hh}\overline{h}_{t-1}+\overline{b}_{hw})\label{eq:rnn_ht}
\end{equation}
$\mathbf W_{hh} \in \R^{D_h \times D_h}$, $\mathbf W_{hw} \in \R^{D_h \times D_w}$, $\overline{b}_{hw} \in \R^{D_h}$, represent the RNN's weights to be optimized and $D_h$, $D_w$ are the dimensions of $\overline{h}_t$ and $\overline{w}_t$ vectors respectively. Note that the weights are shared across the network and $D_w$ could be equal to $K$ the size of the vocabulary (i.e. in case of 1-of-$K$ encoding) or the size of a dense embedding vector generated using a language model such as BERT \cite{devlin2018bert}. The use of RNN allows the model to learn long-range dependencies where the network is unfolded as many times as the length of the sequence (sentence in our case) it is modeling. Although RNNs are capable of handling and representing variable-length sequences, in practice, the learning process faces challenges due to the vanishing/exploding gradient problem \cite{Hochreiter1997,Bengio1994,Graves2012}. In this work, we used gated recurrent unit (GRU) \cite{Cho2014,chung2014empirical} to overcome the latter challenges by updating the computation mechanism of the hidden state vector $\overline{h}_{t}$ through the specified equations below.

\begin{alignat*}{2}
\overline{z}_{t} & =\sigma(\mathbf{W}_{hw}^{z}\overline{w}_{t}+\mathbf{W}_{hh}^{z}\overline{h}_{t-1}+\overline{b}_{hw}^{z}) & \qquad\qquad\qquad\text{(update gate)}\\
\overline{r}_{t} & =\sigma(\mathbf{W}_{hw}^{r}\overline{w}_{t}+\mathbf{W}_{hh}^{r}\overline{h}_{t-1}+\overline{b}_{hw}^{r}) & \qquad\qquad\qquad\text{(reset gate)}\\
\overline{\tilde{h}}_{t} & =\phi(\mathbf{W}_{hw}^{\tilde{h}}\overline{w}_{t}+\overline{r}_{t}\odot\mathbf{W}_{hh}^{\tilde{h}}\overline{h}_{t-1}+\overline{b}_{hw}^{\tilde{h}}) & \qquad\qquad\text{\qquad(new state/memory cell)}\\
\overline{h}_{t} & =(1-\overline{z}_{t})\odot\overline{\tilde{h}}_{t}+z_{t}\odot \overline{h}_{t-1} & \qquad\qquad\qquad\text{(hidden state vector)}
\end{alignat*}
The GRU model computes a reset gate $\overline{r}_t$ that is used to
modulate the effect of the previous hidden state vector $\overline{h}_{t-1}$
when computing the new memory vector $\overline{\tilde{h}}_t$. The
update gate $\overline{z}_t$ determines the importance/contribution
of the newly generated memory vector $\overline{\tilde{h}}_t$ compared
to the previous hidden state vector $\overline{h}_{t-1}$ when computing
the current hidden vector $\overline{h}_t$. The weights $\mathbf{W}^z_{hw}$,
$\mathbf{W}^r_{hw}$, $\mathbf{W}^{\tilde{h}}_{hw}$ each $\in \R^{D_h \times D_w}$
and $\mathbf{W}^z_{hh}$, $\mathbf{W}^r_{hh}$, $\mathbf{W}^{\tilde{h}}_{hh}$
each $\in \R^{D_h \times D_h}$. The biases $\overline{b}^{z}_{hw}$,
$\overline{b}^{r}_{hw}$, $\overline{b}^{\tilde{h}}_{hw}$ each $\in \R^{D_h}$
where $D_h$ and $D_w$ are the dimensions of $\overline{h}_t$ and
$\overline{w}_t$ vectors respectively. The operator $\sigma$
represents the $sigmoid$ function, $\phi$ the $tanh$ or $ReLU$
function, and $\odot$ the element-wise product (i.e. Hadamard product). \newline
The \emph{SentEncoder} uses a bidirectional GRU that computes two hidden state vectors $\overrightarrow{\overline{h}_t}$ and $\overleftarrow{\overline{h}_t}$ for each token $\overline{w}_t$ in sentence $\underline{S}$ corresponding to left-to-right and right-to-left GRU encoding of the sentence. We experimented with two options for computing sentence representation vector $\overline{S}$: (1)\emph{concatenation} $[\overrightarrow{\overline{h}_{T_S}}^\top;\overleftarrow{\overline{h}_{0}}^\top]^\top$, and/or (2) \emph{summation} $[\overrightarrow{\overline{h}_{T_S}} + \overleftarrow{\overline{h}_{0}}]$ of the computed left and right hidden state vectors of the last $\overline{w}_{T_S}$ and first $\overline{w}_{0}$ tokens respectively in sentence $\underline{S}$.
\subsubsection*{Document Encoder (DocEncoder) with Attention}
Originally, each document $\underline{Doc}$ in our corpus is composed of a sequence of sentences (i.e. $\underline{Doc}=[\underline{S}_1, \underline{S}_2, \cdots, \underline{S}_{T_{Doc}}]$ where $\underline{S}_i$ represents the $i^{th}$ sentence and $T_{Doc}$ is the number of sentences in $\underline{Doc}$). Each sentence $\underline{S}_i$ is composed of a sequence of tokens (as described in \emph{SentEncoder} section above) that are processed using \emph{SentEncoder} model to compute the sentence vector representation $\overline{S}_i$. As a result, the processed document $\underline{Doc}^{proc}$ is a sequence of sentences' vector representation (i.e. $\underline{Doc}^{proc}=[\overline{S}_1, \overline{S}_2, \cdots, \overline{S}_{T_{Doc}}]$) that is used as input to \emph{DocEncoder} model.  The \emph{DocEncoder} uses a bidirectional GRU that computes two hidden state vectors $\overrightarrow{\overline{l}_i}$ and $\overleftarrow{\overline{l}_i}$ for each sentence representation $\overline{S}_i$ corresponding to left-to-right and right-to-left GRU encoding of the sentences in $\underline{Doc}^{proc}$. We experimented with two options for joining both hidden state vectors  $\overrightarrow{\overline{l}_i}$ and $\overleftarrow{\overline{l}_i}$ into one vector using: (1)\emph{concatenation} $[\overrightarrow{\overline{l}_i}^\top;\overleftarrow{\overline{l}_i}^\top]^\top$, and/or (2) \emph{summation} $[\overrightarrow{\overline{l}_i} + \overleftarrow{\overline{l}_i}]$ that will be denoted by $\overrightarrow{\overleftarrow{\overline{l}_i}}$ from now on.
Hence, the output of the \emph{DocEncoder} is a sequence of \emph{joined} hidden state vectors $\underline{O}=[\overrightarrow{\overleftarrow{\overline{l}_1}}, \overrightarrow{\overleftarrow{\overline{l}_2}}, \cdots, \overrightarrow{\overleftarrow{\overline{l}_{T_{Doc}}}}]$ that is fed to an \emph{attention} layer to compute the weights associated with each vector which in turn are used to compute a weighted vector sum to obtain a document vector representation $\overline{z}$.

\subsubsection*{Attention Layer}
For many of the DISCERN criteria, pass or fail of the criteria depends on only small fragments throughout the article. For example, for the question \say{Is it clear when the information used or reported in the publication was produced?}, there is likely only one line among a 200+ sentence article (i.e. \say{Last reviewed on...}) that determines whether the article passes the criteria. Our attention-based modeling architecture reflects this problem structure: the model can \say{pay attention} to single sentences within a larger article. We adapt the idea of \emph{global} attention model \cite{Luong2015} in which  a global context/query vector $\overline{q}$ (i.e. trainable parameters in the model) was used along with the output $\underline{O}=[\overrightarrow{\overleftarrow{\overline{l}_1}}, \overrightarrow{\overleftarrow{\overline{l}_2}}, \cdots, \overrightarrow{\overleftarrow{\overline{l}_{T_{Doc}}}}]$ from \emph{DocEncoder} to generate document representation vector $\overline{z}$. The objective is to compute \emph{attention} weights for every $\overrightarrow{\overleftarrow{\overline{l}_i}}$ vector such that $\overline{z}=\sum_{i=1}^{T_{Doc}} \alpha_{i}\overrightarrow{\overleftarrow{\overline{l}_i}}$ where $\alpha_i$ is the normalized weight computed using Eq. \ref{eq:attnw}.
\begin{equation}
\alpha_i = \frac{\exp{(score(\overline{q}, \overrightarrow{\overleftarrow{\overline{l}_i}}))}}{\sum_{k=1}^{T_{Doc}}\exp{(score(\overline{q}, \overrightarrow{\overleftarrow{\overline{l}_k}}))}}\label{eq:attnw}
\end{equation}
For the attention \emph{scoring} function, we experimented with two options inspired by the \emph{additive} approach \cite{Luong2015, Bahdanau2014a} and the \emph{scaled dot-product} work in \cite{Vaswani2017} (see Equations \ref{eq:score_additive} and \ref{eq:score_dotscaled} respectively). In Eq. \ref{eq:score_additive}, the score is computed using three operations: (1) a weight matrix $\mathbf{W}^l_{ql} \in \R^{D_q \times D_l}$ maps $\overleftarrow{\overrightarrow{\overline{l}_i}}$ to a fixed-length vector of dimension equal to the query vector $\overline{q}$ (i.e. $D_q$), (2) a non-linear transformation $tanh$ is applied, and (3) a \emph{dot-product} with $\overline{q}$ is performed. In contrast, in Eq. \ref{eq:score_dotscaled}, the score is computed by performing a \emph{dot-product} between the query vector $\overline{q}$ and $\overrightarrow{\overleftarrow{\overline{l}_i}}$ scaled by $D_l$ which is the dimension of both vectors in similar approach to \cite{Vaswani2017}. Our choice of attention score functions from the vast array of options in the literature \cite{Luong2015, Bahdanau2014a}, was based on limiting the number of parameters in our model given the size of our dataset. 
\begin{equation}
score(\overline{q}, \overrightarrow{\overleftarrow{\overline{l}_i}}) = \overline{q}^\top tanh( \mathbf{W}^l_{ql}\overrightarrow{\overleftarrow{\overline{l}_i}})\label{eq:score_additive}
\end{equation}
\begin{equation}
score(\overline{q}, \overrightarrow{\overleftarrow{\overline{l}_i}}) = \frac{\overline{q}^\top \overrightarrow{\overleftarrow{\overline{l}_i}}}{\sqrt{D_l}}\label{eq:score_dotscaled}
\end{equation}

\subsubsection*{Decoder/output classifier}

The last layer in the HEA model takes as input the computed document representation vector $\overline{z}$ from the \emph{DocEncoder} layer and performs an affine transformation followed by \emph{softmax} operation to compute a probability distribution on the labels for the document under consideration.
That is, the outcome $\hat{y}$ for a given Brief DISCERN criterion is computed using Eq. \ref{eq:y_hat}
\begin{equation}
\hat{y}=\sigma(\mathbf{W}_{V_{label}z}\overline{z}+\overline{b}_{V_{label}})\label{eq:y_hat}
\end{equation}
where $\mathbf W_{V_{label}z} \in \R^{|V_{label}| \times D_z}$, $\overline{b}_{V_{label}} \in \R^{|V_{label}|}$ represents the classifier's weights to be optimized, $V_{label} \in \{0,1\}$ is the set of admissible labels for a criterion (binary variable in our case), $|V_{label}|$ is the number of labels, $D_z$ is the dimension of $\overline{z}$ (document representation vector), and $\sigma$ is the $softmax$ function. As a result, the outcome $\hat{y}$ represents a probability distribution over the set of possible labels $V_{label}$.
\subsubsection*{Objective function}
We used cross-entropy loss as our objective function for each Brief DISCERN criterion model. The loss function for a $j^{th}$ document is defined by Eq. \ref{eq:loss_doc} where $y_{c}\in\{0,1\}$ is equivalent to $\mathbbm{1}{\big[y = c\big]}$ (i.e. a boolean indicator equal to 1 when $c$ is the reference/ground-truth class), and $\hat{y}_{c}$ is the probability of the class $c$. The objective function for the whole training set $D_{train}$ is defined by the average loss across all the documents in $D_{train}$ plus a weight regularization term (i.e. $l_2$-norm regularization) applied to the model parameters represented by $\bm{\theta}$ (see Eq. \ref{eq:loss_corpus}).
\begin{equation}
l^{(j)}=-\sum_{c=1}^{|V_{label}|}y_{c}^{(j)}\times log(\hat{y}_{c}^{(j)}) \label{eq:loss_doc}
\end{equation}
\begin{equation}
L(\mathbf{\bm{\theta}}) =\frac{1}{N}\sum_{j=1}^{N}l_{j} + \frac{\lambda}{2}||\mathbb{\bm{\theta}}||_{2}^{2}\label{eq:loss_corpus}
\end{equation}
In addition to the $l_{2}$-norm regularization, we also experimented with dropout \cite{Srivastava2014} by deactivating neurons in the network layers using probability $p_{dropoout}$. Moreover, we used pre-trained language models such as BERT \cite{devlin2018bert, Wolf2019} and BioBERT \cite{10.1093/bioinformatics/btz682} to extract token embeddings that are used as input to HEA's model (i.e. representation of token $\overline{w}_t$).

We additionally implemented a neural-based model (HE) that follows the same architecture of HEA model but without the attention layer such that the output of the\emph{DocEncoder} representing a sequence of \emph{joined} hidden state vectors $\underline{O}=[\overrightarrow{\overleftarrow{\overline{l}_1}}, \overrightarrow{\overleftarrow{\overline{l}_2}}, \cdots, \overrightarrow{\overleftarrow{\overline{l}_{T_{Doc}}}}]$ is mean pooled (i.e. averaged) to obtain an overall document vector representation $\overline{z}$.

\subsection*{Hyperparameter optimization for neural models}

We developed a multiprocessing module that used a uniform random search strategy \cite{BergstraJAMESBERGSTRA2012} that randomly chose a set of hyperparameters configurations (i.e. layer depth, embedding size, attention approach, etc.) from the set of all possible configurations. Then the best configuration for each model (i.e. the one achieving best performance on the validation set) was used for the final training and testing.

\subsection*{Baseline Machine Learning Models}
For the traditional modeling approach, the content of each article was converted into a bag of words representation and weighted using the term frequency–inverse document frequency (TF-IDF) weighting scheme. We also computed a set of features based on the existence of HTML links, bibliography keywords, references to medical terms (extracted using MetaMap Lite \cite{Demner-Fushman2017}), and named entities within the text, as well as a measure of text polarity. Recursive feature elimination with cross validation was used to identify the optimal subset of features. For its ease of interpretability and good performance on feature sets with many categorical variables, we implemented a Random Forest model with scikit-learn \cite{scikit-learn} to predict if the criterion is fulfilled or not for every criterion in Brief DISCERN.

\subsection*{Experimental setup}

We followed a stratified 5-fold cross-validation scheme where each fold was defined as a distinct 80\%-20\% train-test split. Due to the imbalance in outcome classes, training examples were weighted inversely proportional to class/outcome frequencies in the training data. Articles from the three health topics were randomly distributed between the 5 folds. Within each fold, parameter selection was performed with a validation set consisting of 10\% of the training set. During the training of the models, the epoch in which the model achieved the best F1-macro score on the validation set was recorded, and model state as it was trained up to that epoch was saved. This best model, as determined by the validation set, was then tested on the test split.

Model performance was evaluated using F1-macro and classification accuracy. In this quality assessment problem, we value precision equally with recall, so F1 is a good measure that captures both. The evaluation of the trained models was based on their average performance on the test sets of the five folds.

We also performed a coverage analysis to determine how the model could be adapted to handle uncertainty. In addition to classifying articles as low or high quality, we also have the option of allowing the model report that it is unsure about a criteria. In instances when the model has a low confidence in its prediction, it is more valuable to the user for the model to convey that uncertainty, than to make a less accurate prediction. In addition, there is also the option to send articles where the model has low confidence to a human for manual evaluation. However, there is a direct trade-off between the quality (accuracy) and the quantity the predictions; by requiring a higher threshold of confidence, the model will by definition make a fewer number of predictions. The frequency with which the model makes prediction above a certain confidence threshold, i.e. outputs a prediction to the user, is called coverage. We calculated the models' accuracy at different levels of coverage and their associated confidence thresholds. For example, to calculate the accuracy associated with a coverage of 80\%, we computed the 20th percentile prediction confidence score, and computed accuracy metrics on only the articles with prediction confidence scores (i.e. the probability from \emph{softmax} layer) that exceed the 20th percentile. Predictions that are below this threshold are considered \emph{unsure}. These are instances where the model would abstain from making a prediction, or the article could be sent for manual review.

\subsection*{Code Availability}

The data preprocessing and the models' implementation (training and testing) workflow is made publicly available at \url{https://github.com/uzh-dqbm-cmi/auto-discern}. 

\section*{Results}

\begin{figure}
  \centering
      \includegraphics[width=\textwidth]{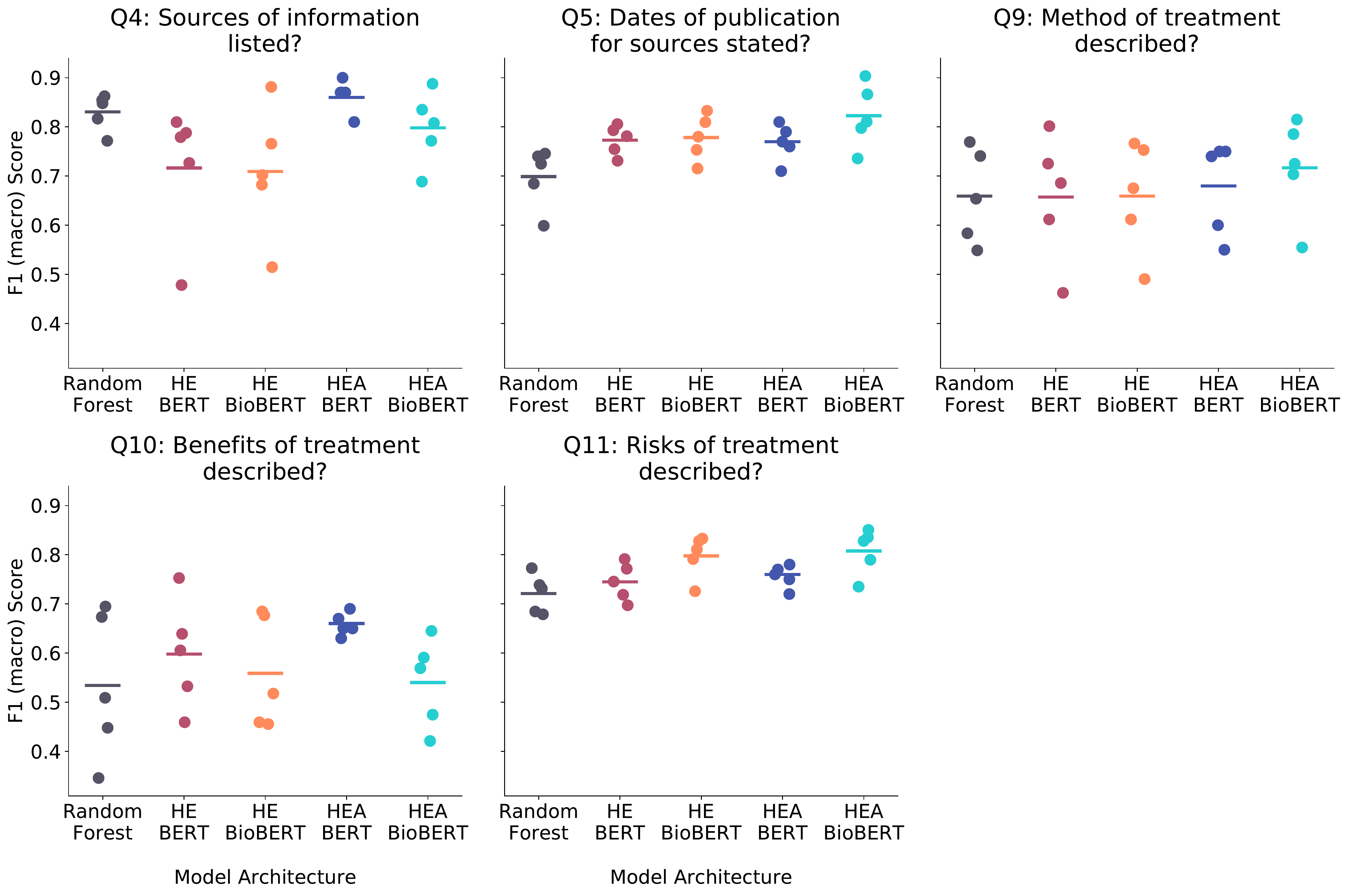}
  \caption{Performance comparison of the model architectures on each of the Brief-DISCERN questions. Each point represents the performance of the architecture on each of the 5 cross validation folds.}
  \label{fig:modelsperf_f1macro}
\end{figure}

\begin{table}
    \caption{Average F1-macro scores with standard deviation by model architecture.}
   \label{table:f1macro}
  \begin{center}
        \normalsize
        \begin{tabular}
          {@{} 
            l
            >{}S[table-format = 2.2(2)]
            S[table-format = 2.2(2)]
            S[table-format = 2.2(2)]
          @{}}
          \toprule
          {Model Architecture}  &  {Q4: References} & {Q5: Date} & {Q9: How Treatment Works} \\
          \midrule
            Random Forest   &  0.83 ( 4) &  0.70 ( 6) &  0.66 (10) \\
            HE BERT         &  0.72 (14) &  0.77 ( 3) &  0.66 (13) \\
            HE BioBERT      &  0.71 (13) &  0.78 ( 5) &  0.66 (11) \\
            HEA BERT        &  0.86 ( 3) &  0.77 ( 4) &  0.68 (10) \\
            HEA BioBERT     &  0.80 ( 7) &  0.82 ( 6) &  0.72 (10) \\
          \midrule
          \\
        \end{tabular}
        
        \begin{tabular}
          {@{} 
            l
            >{}S[table-format = 2.2(2)]
            S[table-format = 2.2(2)]
            S[table-format = 2.2(2)]
          @{}}
          {Model Architecture}  &  {Q10: Tt. Benefits} & {Q11: Tt. Risks} & {All Questions Avg}   \\
          \midrule
            Random Forest   &  0.53 (15) &  0.72 ( 4)  & 0.69 \\
            HE BERT         &  0.60 (11) &  0.74 ( 4)  & 0.70 \\
            HE BioBERT      &  0.56 (11) &  0.80 ( 4)  & 0.70 \\
            HEA BERT        &  0.66 ( 2) &  0.76 ( 3)  & 0.75 \\
            HEA BioBERT     &  0.54 ( 9) &  0.81 ( 5)  & 0.74 \\
          \bottomrule
        \end{tabular}
    \end{center}
\end{table}

We compared the performance of the five trained models (Random Forest, HEA with BERT and BioBERT embeddings, and HE with BERT and BioBert embeddings) across all five folds using F1-macro scores (Table \ref{table:f1macro} and Figure \ref{fig:modelsperf_f1macro}). Overall, the HEA architecture perforemd the best, scoring an average F1-macro score of 0.75 with BERT embeddings and 0.74 with BioBERT embeddings. In comparison, the HE architectures without the attention mechanism averaged 0.70 on both embeddings. The Random Forest model achieved an average F1-macro score of 0.69.

Almost all models performed the best on question 4 (\say{Is it clear what sources of information were used to compile the publication [other than the author or producer]?}) with HEA BERT, HEA BioBERT and Random Forest scoring 0.86, 0.80, and 0.83 respectively. The HE models performed worse on this question, with 0.72 with BERT and 0.71 with BioBERT. All five models achieved high F1-macro scores on question 5 (\say{Is it clear when the information used or reported in the publication was produced?}) with HEA BioBERT coming first (0.82), HE BioBERT second (0.78), HEA BERT and HE BERT tying for third (0.77), and Random Forest last (0.70). 

For treatment related questions, HEA BioBert performed the best. On question 9 (\say{Does it describe how each treatment works?}), HEA BioBERT came first with an average F1-macro score of 0.72, and the remaining models ranging between 0.68 and 0.66). For question 11 (\say{Does it describe the risks of each treatment?}), both neural models using the BioBERT Embeddings performed the best: the HEA- and HE BioBERT models scored 0.81 and 0.80 respectively, with the remaining models following betwwen 0.76 and 0.72. In contrast, for question 10 (\say{Does it describe the benefits of each treatment?}), the neural models using the BERT embeddings performed better, with HEA BERT at 0.66 and HE BERT at 0.60, with the remaining models ranging between 0.56 and 0.53. It worth mentioning that Q10 has the greatest class imbalance (i.e. 77\% of the articles in the data set described the benefits of the treatment).

\begin{figure}
  \centering
      \includegraphics[width=12cm]{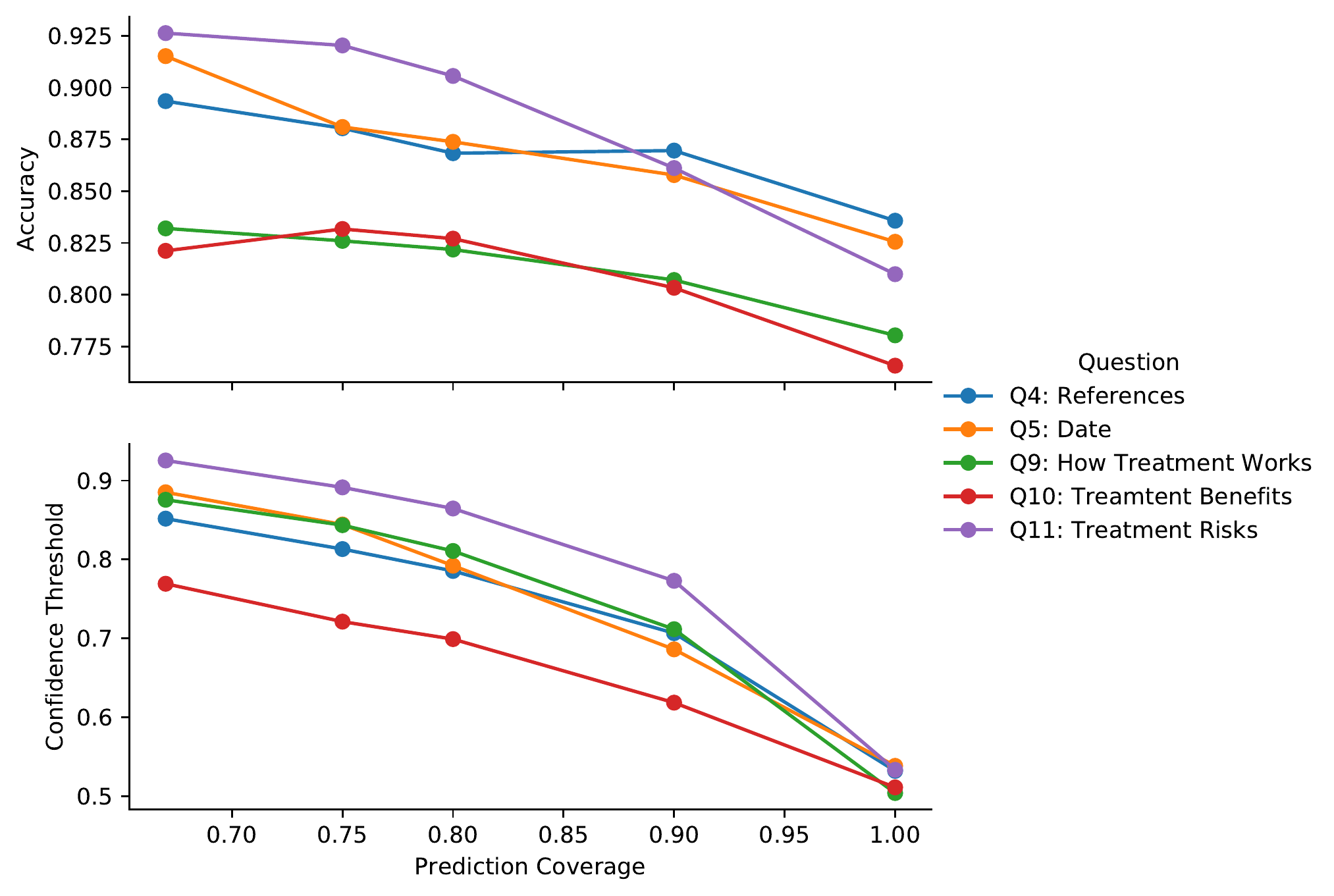}
  \caption{Relationship between Prediction Coverage, Confidence Threshold, and Model Accuracy.}
  \label{fig:coverage_threshold_accuracy}
\end{figure}

\begin{table}[ht]
\caption{Comparison of performance metrics for the HEA BioBERT architecture at 80\% and 100\% coverage. Coverage refers to the percent of articles the model makes a prediction for (as opposed to abstaining from making a prediction when the model has a confidence below the Threshold). The Precision, Recall, and Accuracy scores reflect the accuracy of the model on the resulting 80\% of predicted articles. }
\label{implementation_metrics}
\begin{center}
    \normalsize
      \begin{tabular}{lccccc}
      \toprule
        Question                & Coverage  & Threshold & Precision & Recall & Accuracy \\ 
        \midrule
        Q4: References          & 80\%      & 0.79      & 0.87      & 0.79   & 87\% \\
        Q5: Date                & 80\%      & 0.79      & 0.87      & 0.88   & 87\% \\
        Q9: How Treatment Works & 80\%      & 0.81      & 0.84      & 0.71   & 82\% \\
        Q10: Treatment Benefits & 80\%      & 0.70      & 0.66      & 0.55   & 83\% \\
        Q11: Treatment Risks    & 80\%      & 0.86      & 0.90      & 0.90   & 91\% \\
        \midrule
        Q4: References          & 100\%     & 0.50      & 0.83      & 0.80   & 84\% \\
        Q5: Date                & 100\%     & 0.50      & 0.83      & 0.83   & 83\% \\
        Q9: How Treatment Works & 100\%     & 0.50      & 0.77      & 0.72   & 78\% \\
        Q10: Treatment Benefits & 100\%     & 0.50      & 0.57      & 0.54   & 77\% \\
        Q11: Treatment Risks    & 100\%     & 0.50      & 0.81      & 0.81   & 81\% \\
        \bottomrule
      \end{tabular}
      \end{center}
\end{table}

As measured by F1-macro, HEA-BioBert took first place in 3 of the 5 questions, and HEA-Bert took first in the remaining 2 questions (see Table \ref{table:f1macro}). However, when computing the average score on all questions, HEA BERT and BioBERT performed comparably, and the variance across folds was lower for HEA BERT compared to HEA BioBERT (see Figure \ref{fig:modelsperf_f1macro}).

We explored the relationship between model's coverage, accuracy, and confidence (prediction probability threshold) focusing on the BioBERT model (see Figure \ref{fig:coverage_threshold_accuracy}). The trend is that as the model's coverage decreases, the higher is the confidence (i.e. outcome probability) and the accuracy of the model. At 80\% coverage, the model achieves 86\% average accuracy with average confidence equal to 0.79 (see Table \ref{implementation_metrics}). 

Table \ref{table:humanvsmachine} compares the machine learning model (HEA BioBERT) performance to human manual performance on the DISCERN and HON guidelines. 
We report the DISCERN Manual Performance as the frequency with which each of our two raters' agree with the aggregated average of both raters (i.e. percent agreement to aggregate).   The manual rater accuracy score  on  our  data  set  averaged  94\% (spanning 88\% - 97\% across criteria). Compared to the DISCERN raters' manual performance of 94\%, the models performance was adequate (81\% accuracy across all questions at 100\% coverage). In order to bring the model's performance closer to the DISCERN raters' manual performance, we could reduce the model coverage to 80\%, which would yield an accuracy of 86\%.

As an additional comparison, Table \ref{table:humanvsmachine} also contains HON organization manual percent agreement scores that were computed while developing training sets for their own machine learning models \cite{Boyer2015}. The average percent agreement among the HON raters was 85\% on their full criteria, and this drops to 81\% when only considering criteria that are shared with Brief DISCERN (Reference, Date, and Justifiability). Overall, the machine learning model achieved a competitive performance at full coverage compared to HON raters: the model averaged 83\% accuracy at 100\% coverage on the questions that overlap with HON, and the HON raters had percent agreement of 81\%. 

Table \ref{table:attn_sent} shows top-3 sentences (based on attention probability score) belonging to most confidently predicted documents as determined by the prediction probability score for each question in the three medical topics.

Lastly, the time and space requirements for the different architectures were very different. For the neural models, running the full hyperparameter search and training routine took between 25-30 hours on a GPU node with 256 GB of RAM parallelized across 5 Nvidia GTX 1080 GPUs. In comparison, the baseline model trained in 15 minutes on a machine with 4 CPUs and 16 GB of RAM.

\begin{table}[h!]
    \caption{Performance Comparison between Human Manual Rating and Deep Learning Model. Manual performance is reported as interrater agreement. Automated performance is reported as Implementation Accuracy (see Table \ref{implementation_metrics}).}
    \label{table:humanvsmachine}
    \begin{center}
        \normalsize
        \begin{tabular}{@{} l | c c | c c @{}}
          \toprule
            \multicolumn{1}{l}{Question} & \multicolumn{2}{c}{Manual Performance} & \multicolumn{2}{c}{Automated Performance} \\
                                           & \multicolumn{1}{c}{DISCERN} & {HONcode}     & \multicolumn{2}{c}{DISCERN HEA BioBERT} \\
                                           & \multicolumn{1}{c}{2 raters} & {3 raters}     & \multicolumn{1}{c}{80\% coverage} & {100\% coverage} \\
          \midrule
            Q4: References (HoN: Reference)& 96\%      &  89\%          &  87\% & 84\%\\ 
            Q5: Date (HoN: Date)           & 88\%      &  80\%          &  87\% & 83\%\\
            Q9: How Treatment Works        & 92\%      &                &  82\% & 78\%\\
            Q10: Treatment Benefits        & 95\%      &                &  83\% & 77\% \\
            Q11: Treatment Risks (HoN: Justifiability) & 97\%      &  74\%          &  91\% & 81\%\\
            \midrule 
            average                        & 94\%      & 81\%           & 86\%  & 81\%\\
          \bottomrule
        \end{tabular}
    \end{center}
\end{table}

\begin{table}[hb]
    \caption{Example sentences that the models paid the most attention to for each disease category. These are the sentences with the highest attention weight for the top three most confidently predicted documents as determined by the prediction probability score. These results are from the HEA BioBERT model.}
    \label{table:attn_sent}
    \begin{center}
    \begin{tabular}{@{} p{5.5cm} | p{6.5cm} | p{5.5cm} @{}}
        \multicolumn{3}{l}{\textbf{Question 4: Sources}} \\
            \toprule
            \multicolumn{1}{l}{Breast Cancer} & \multicolumn{1}{l}{Arthritis}      & \multicolumn{1}{l}{Depression} \\
            \midrule
                2010 Aug 10;28(23):3784-96.   & Nat Rev Rheumatol.                 & J Abnorm Psychol. \\  
                2008;148(5): 358-69.          & Leuk Res.                          & J Abnorm Psychol. \\  
                Lancet 2007; 369(9555):29–36. & Kelley's Textbook of Rheumatology. & American Journal of Geriatric Psychiatry. \\ 

        \multicolumn{3}{l}{\textbf{Question 5: Date}} \\
            \toprule
            \multicolumn{1}{l}{Breast Cancer} & \multicolumn{1}{l}{Arthritis}      & \multicolumn{1}{l}{Depression} \\
            \midrule
                Review Date: 11/17/2012.      & Review Date: 9/26/2011.            & Review Date: 3/8/2013.         \\
                Last Revised: 10/01/2013.     & All rights reserved.               & All rights reserved.           \\
                Review Date: 6/5/2012.        & Article updated: 31 October 2012.  & Page last updated: 1-Oct-2013. \\

        \multicolumn{3}{l}{\textbf{Question 9: How Treatment Works}} \\
            \toprule
            \multicolumn{1}{l}{Breast Cancer} & \multicolumn{1}{l}{Arthritis}      & \multicolumn{1}{l}{Depression} \\
            \midrule
                
                During this surgery, the surgeon removes the axillary lymph nodes as well as the chest wall muscle in addition to the breast. & 
                In this surgery, the healthcare provider actually removes the inflamed synovial tissue. &
                The basis of this therapy is that behaviours such as inactivity and ruminating on certain thoughts can be key factors in maintaining  depression.  \\
                
                Radiation therapy is typically done using a large machine that aims the energy beams at your body (external beam radiation). & 
                One part of such therapy involves working with a physical therapist to perform dedicated exercises for muscle strengthening, increasing range  &
                Gentler martial arts which focus on internal control, breathing and mental discipline can be especially useful for combating depressed  thinking and improving relaxation skills.  \\
                
                Three-dimensional conformal radiation therapy (3D-CRT): As part of this treatment, special computers create detailed three-dimensional pictures & 
                Hydrotherapy differs from swimming because it involves special exercises that you do in a warm-water pool.  & 
                Psychoanalytic therapists rely on suggestion, hypnosis, and reeducation to reform self-esteem, and helps the person construct coping  strategies to deal with grief, sadness, disappointment, achievement, and pleasure.  \\

        \multicolumn{3}{l}{\textbf{Question 10: Treatment Benefits}} \\
            \toprule
            \multicolumn{1}{l}{Breast Cancer} & \multicolumn{1}{l}{Arthritis}      & \multicolumn{1}{l}{Depression} \\
            \midrule
                
                Treating early breast cancer.  & 
                Getting Established on DMARD Therapy.  &
                Cognitive Behavioral Therapy for Depression.   \\

                Targeted therapy for breast cancer.  & 
                Medications will not JIA; rather they can help to symptoms and keep disease activity under .  &
                The mindfulness approach uses meditation, yoga, and breathing exercises to focus awareness on the present moment and break negative thinking    \\

                Adjuvant and Neoadjuvant Therapy for Breast Cancer.  & 
                Treatment for Juvenile Rheumatoid Arthritis.  &
                CBT is based on two specific tasks: cognitive restructuring, in which the therapist and patient work together to change thinking patterns,  192 and  behavioral activation -- in which patients learn to overcome obstacles to participating in enjoyable activities.   \\

        \multicolumn{3}{l}{\textbf{Question 11: Treatment Risks}} \\
            \toprule
            \multicolumn{1}{l}{Breast Cancer} & \multicolumn{1}{l}{Arthritis}      & \multicolumn{1}{l}{Depression} \\
            \midrule
                
                The side effects vary depending on which biological therapy drug you have.  & 
                Risks: Always talk to your doctor or pharmacist before taking NSAIDs as they may cause serious side effects compared to paracetamol.  &
                Side Effects of ECT .   \\

                Side effects .  & 
                Risks: Always talk to your doctor or pharmacist before taking NSAIDs as they may cause serious side effects compared to paracetamol.  &
                Common side effects of SSRIs include:.   \\

                Are there side effects or risks from hormone therapy?  & 
                Common side effects include a rise in blood pressure, increased hair growth, increased swelling of the gums and an increased risk of  developing an infection.  &
                What Are the Risks?   \\    
        \bottomrule
        \end{tabular}
    \end{center}
\end{table}

\section*{Discussion}

In this research, we developed an attention-based neural network model with the aim to automatically determine the quality of online health information. 

The experiments suggest that a neural network model with trained language embeddings on large text corpora (generic or medical) has better performance than a conventional baseline model (Random Forest). Importantly, this superior performance was achieved without the need to hand-craft input features, as was the case with the baseline model. However, it is worth noting that this comes at the trade-off of much higher computing requirements for the neural network models.

Our results reiterate the success of using trained language models \cite{Vaswani2017, Wolf2019}, and transfer learning \cite{Ruder2019} in achieving competitive results even on small datasets (as in our case). The BioBERT embeddings show a slight advantage in comparison to BERT ones (Table \ref{table:f1macro}), and we believe this could be due to the medical topics and the language used to describe treatments in each topic. 

Our results suggest that the neural attention mechanism not only provided a performance boost over a mean pooled neural architecture, but also enabled greater model explainability. The HEA models performed 7\% higher in F1-macro compared to the HE models (Table \ref{table:f1macro}).
In addition, inspecting Table \ref{table:attn_sent}, it can be seen that the HEA BioBERT model provided reasonable context sentences (i.e. sentences supporting a prediction). In other words, the models identify textual snippets (surrogates, or proxy) in the articles. For question 4 (References), the model identified sentences containing citations, and in question 5 (Date), the model identified text referring to dates when the article was \say{reviewed}, \say{revised}, or \say{updated}. 
Questions 9 (How Treatment Works) and 10 (Treatment Benefits) are related questions, and we see that the textual snippets identified by the model for these questions overlap, as expected. However, question 10 achieved poorer accuracy scores, which is probably due to class imbalance in the training data for that question (only 33\% of articles were in the negative class). 
For question 11 (Treatment Risks), the model often identified section headings containing the phrase \say{side effects}. 

We compared the models' quality assessment performance to  humans manually performing the same task. Our raters achieved an average of 94\% agreement across criteria. Similarly, the HON organization reported an average percent agreement of 85\% on their criteria, and this drops to 81\% when only considering criteria that are shared with Brief DISCERN. While the HEA BioBERT model performed lower than manual raters used in this study (81\% vs. 94\%), it showed competitive results when compared to HON raters (83 \% vs. 81\% average accuracy). Restricting the HEA BioBERT model's prediction coverage to 80\%, we could further improve the prediction performance achieving 86\% average accuracy across all criteria. In this case, the model refrains from making a prediction when its prediction probability score, or confidence, is below the 20th percentile. This model or a similarly trained model could be effectively used for pre-screening health web pages and for assisting manual raters in the quality assessment task. As suggested by the HON organization, assisting manual rating with automated systems could reduce manual effort \cite{Boyer2015}.

 \subsection*{Future Work}

We are seeking to further improve our models' performance to more closely achieve human performance in assessing online health information quality. One straightforward approach is to train on a larger data set using the same model architecture. To achieve this aim, we could look beyond manually labeling more health articles. For example, we could construct a larger corpora by combining our current dataset with other existing bodies of online health information that have independently been assessed as being of high quality. For example, articles approved by HON could be used as positive examples in an augmented training set. An additional avenue is to use semi-supervised learning and unsupervised data augmentation approaches \cite{Berthelot2019,Xie2019} where unlabeled data is incorporated to improve classification performance without additional annotation burden.

In future experiments, we plan to further develop our use of language embeddings. For example, in this research we simply used the last layer embeddings from the BERT and BioBERT models. However, recent experiments suggest using different layers (the BERT network contains 12 layers) or further training the embedding networks can yield performance improvements. 

Finally, the DISCERN instrument is designed to be applied to articles describing treatment options. Thus, our model's applicability is limited to these types of articles. Similarly, our model does not extend to the medium of social media, which online users are increasingly using to share and consume health information \cite{Viviani2017}. More research is needed to develop models for assessing the quality of other types and mediums of health information. 

\section*{Conclusion}
Our study demonstrates that neural models are able to perform online health information quality assessment in accordance with an existing quality criteria (Brief DISCERN) with a performance above 80\% accuracy. The neural approach achieves a better performance than a conventional approach using Random Forest. In addition, we observe that existing biomedical language models improve performance on this task. Finally, we show that attention-based neural approaches are able to retrieve relevant supporting sentences from the text, which makes model decisions more explainable to users. 


\subsection*{Availability of data and materials}
The preprocessing and the models’ implementation (training and testing) workflow is made publicly available at \url{https://github.com/uzh-dqbm-cmi/auto-discern}

\subsection*{Competing interests}
    The authors declare that they have no competing interests.

\subsection*{Acknowledgements}
The authors would like to acknowledge the Swiss National Science Foundation for their previous funding of A-Discern project (grant number P2TIP1-161635) awarded to AA.

\subsection*{Author's contributions}
LK and AA worked on the development of processing and analysis workflow, algorithms and models implementation. LK, AA and MK analyzed and interpreted the data. LK and AA drafted the manuscript. AA and MK supervised and edited the manuscript. All authors approved the final article.

\bibliographystyle{bmc-mathphys} 
\bibliography{article.bib}      


\begin{thebibliography}{37}
\ifx \bisbn   \undefined \def \bisbn  #1{ISBN #1}\fi
\ifx \binits  \undefined \def \binits#1{#1}\fi
\ifx \bauthor  \undefined \def \bauthor#1{#1}\fi
\ifx \batitle  \undefined \def \batitle#1{#1}\fi
\ifx \bjtitle  \undefined \def \bjtitle#1{#1}\fi
\ifx \bvolume  \undefined \def \bvolume#1{\textbf{#1}}\fi
\ifx \byear  \undefined \def \byear#1{#1}\fi
\ifx \bissue  \undefined \def \bissue#1{#1}\fi
\ifx \bfpage  \undefined \def \bfpage#1{#1}\fi
\ifx \blpage  \undefined \def \blpage #1{#1}\fi
\ifx \burl  \undefined \def \burl#1{\textsf{#1}}\fi
\ifx \doiurl  \undefined \def \doiurl#1{\textsf{#1}}\fi
\ifx \betal  \undefined \def \betal{\textit{et al.}}\fi
\ifx \binstitute  \undefined \def \binstitute#1{#1}\fi
\ifx \binstitutionaled  \undefined \def \binstitutionaled#1{#1}\fi
\ifx \bctitle  \undefined \def \bctitle#1{#1}\fi
\ifx \beditor  \undefined \def \beditor#1{#1}\fi
\ifx \bpublisher  \undefined \def \bpublisher#1{#1}\fi
\ifx \bbtitle  \undefined \def \bbtitle#1{#1}\fi
\ifx \bedition  \undefined \def \bedition#1{#1}\fi
\ifx \bseriesno  \undefined \def \bseriesno#1{#1}\fi
\ifx \blocation  \undefined \def \blocation#1{#1}\fi
\ifx \bsertitle  \undefined \def \bsertitle#1{#1}\fi
\ifx \bsnm \undefined \def \bsnm#1{#1}\fi
\ifx \bsuffix \undefined \def \bsuffix#1{#1}\fi
\ifx \bparticle \undefined \def \bparticle#1{#1}\fi
\ifx \barticle \undefined \def \barticle#1{#1}\fi
\ifx \bconfdate \undefined \def \bconfdate #1{#1}\fi
\ifx \botherref \undefined \def \botherref #1{#1}\fi
\ifx \url \undefined \def \url#1{\textsf{#1}}\fi
\ifx \bchapter \undefined \def \bchapter#1{#1}\fi
\ifx \bbook \undefined \def \bbook#1{#1}\fi
\ifx \bcomment \undefined \def \bcomment#1{#1}\fi
\ifx \oauthor \undefined \def \oauthor#1{#1}\fi
\ifx \citeauthoryear \undefined \def \citeauthoryear#1{#1}\fi
\ifx \endbibitem  \undefined \def \endbibitem {}\fi
\ifx \bconflocation  \undefined \def \bconflocation#1{#1}\fi
\ifx \arxivurl  \undefined \def \arxivurl#1{\textsf{#1}}\fi
\csname PreBibitemsHook\endcsname

\bibitem{Hesse2005}
\begin{barticle}
\bauthor{\bsnm{Hesse}, \binits{B.W.}},
\bauthor{\bsnm{Nelson}, \binits{D.E.}},
\bauthor{\bsnm{Kreps}, \binits{G.L.}},
\bauthor{\bsnm{Croyle}, \binits{R.T.}},
\bauthor{\bsnm{Arora}, \binits{N.K.}},
\bauthor{\bsnm{Rimer}, \binits{B.K.}},
\bauthor{\bsnm{Viswanath}, \binits{K.}}:
\batitle{{Trust and Sources of Health Information}}.
\bjtitle{Archives of Internal Medicine}
\bvolume{165}(\bissue{22}),
\bfpage{2618}
(\byear{2005}).
doi:\doiurl{10.1001/archinte.165.22.2618}
\end{barticle}
\endbibitem

\bibitem{Fahy2014}
\begin{barticle}
\bauthor{\bsnm{Fahy}, \binits{E.}}:
\batitle{{Quality of patient health information on the internet: reviewing a
  complex and}}.
\bjtitle{Australasian Medical Journal}
\bvolume{7}(\bissue{1}),
\bfpage{24}--\blpage{28}
(\byear{2014}).
doi:\doiurl{10.4066/AMJ.2014.1900}
\end{barticle}
\endbibitem

\bibitem{Zhang2015}
\begin{botherref}
\oauthor{\bsnm{Zhang}, \binits{Y.}},
\oauthor{\bsnm{Sun}, \binits{Y.}},
\oauthor{\bsnm{Xie}, \binits{B.}}:
{Quality of health information for consumers on the web: A systematic review of
  indicators, criteria, tools, and evaluation results}.
Journal of the Association for Information Science and Technology
\textbf{66}(10)
(2015).
doi:\doiurl{10.1002/asi.23311}
\end{botherref}
\endbibitem

\bibitem{Saunders2018}
\begin{botherref}
\oauthor{\bsnm{Saunders}, \binits{C.H.}},
\oauthor{\bsnm{Petersen}, \binits{C.L.}},
\oauthor{\bsnm{Durang}, \binits{M.-A.}},
\oauthor{\bsnm{Bagley}, \binits{P.J.}},
\oauthor{\bsnm{Elywn}, \binits{G.}}:
{Bring on the Machines: Could Machine Learning Improve the Quality of Patient
  Education Materials? A Systematic Search and Rapid Review}.
American Society of Clinical Oncology
(2018)
\end{botherref}
\endbibitem

\bibitem{Murray2003}
\begin{barticle}
\bauthor{\bsnm{Murray}, \binits{E.}},
\bauthor{\bsnm{Lo}, \binits{B.}},
\bauthor{\bsnm{Pollack}, \binits{L.}},
\bauthor{\bsnm{Donelan}, \binits{K.}},
\bauthor{\bsnm{Catania}, \binits{J.}},
\bauthor{\bsnm{Lee}, \binits{K.}},
\bauthor{\bsnm{Zapert}, \binits{K.}},
\bauthor{\bsnm{Turner}, \binits{R.}}:
\batitle{{The Impact of Health Information on the Internet on Health Care and
  the Physician-Patient Relationship: National U.S. Survey among 1.050 U.S.
  Physicians}}.
\bjtitle{Journal of Medical Internet Research}
\bvolume{5}(\bissue{3}),
\bfpage{17}
(\byear{2003}).
doi:\doiurl{10.2196/jmir.5.3.e17}
\end{barticle}
\endbibitem

\bibitem{Allam2014}
\begin{barticle}
\bauthor{\bsnm{Allam}, \binits{A.}},
\bauthor{\bsnm{Schulz}, \binits{P.J.}},
\bauthor{\bsnm{Nakamoto}, \binits{K.}}:
\batitle{{The impact of search engine selection and sorting criteria on
  vaccination beliefs and attitudes: two experiments manipulating Google
  output.}}
\bjtitle{Journal of medical Internet research}
\bvolume{16}(\bissue{4}),
\bfpage{100}
(\byear{2014}).
doi:\doiurl{10.2196/jmir.2642}
\end{barticle}
\endbibitem

\bibitem{Ludolph2016}
\begin{barticle}
\bauthor{\bsnm{Ludolph}, \binits{R.}},
\bauthor{\bsnm{Allam}, \binits{A.}},
\bauthor{\bsnm{Schulz}, \binits{P.J.}}:
\batitle{{Manipulating Google's Knowledge Graph Box to Counter Biased
  Information Processing During an Online Search on Vaccination: Application of
  a Technological Debiasing Strategy}}.
\bjtitle{Journal of Medical Internet Research}
\bvolume{18}(\bissue{6}),
\bfpage{137}
(\byear{2016}).
doi:\doiurl{10.2196/jmir.5430}
\end{barticle}
\endbibitem

\bibitem{Iverson2008}
\begin{barticle}
\bauthor{\bsnm{Iverson}, \binits{S.A.}},
\bauthor{\bsnm{Howard}, \binits{K.B.}},
\bauthor{\bsnm{Penney}, \binits{B.K.}}:
\batitle{Impact of internet use on health-related behaviors and the
  patient-physician relationship: a survey-based study and review.}
\bjtitle{The Journal of the American Osteopathic Association}
\bvolume{108}(\bissue{12}),
\bfpage{699}--\blpage{711}
(\byear{2008})
\end{barticle}
\endbibitem

\bibitem{Wald2007}
\begin{barticle}
\bauthor{\bsnm{Wald}, \binits{H.S.}},
\bauthor{\bsnm{Dube}, \binits{C.E.}},
\bauthor{\bsnm{Anthony}, \binits{D.C.}}:
\batitle{{Untangling the Web-The impact of Internet use on health care and the
  physician-patient relationship}}.
\bjtitle{Patient Education and Counseling}
\bvolume{68}(\bissue{3}),
\bfpage{218}--\blpage{224}
(\byear{2007}).
doi:\doiurl{10.1016/j.pec.2007.05.016}
\end{barticle}
\endbibitem

\bibitem{Risk2001}
\begin{botherref}
\oauthor{\bsnm{Risk}, \binits{A.}},
\oauthor{\bsnm{Dzenowagis}, \binits{J.}}:
{Review of Internet health information quality initiatives}.
Journal of Medical Internet Research
(2001).
doi:\doiurl{10.2196/jmir.3.4.e28}.
\url{http://www.jmir.org/2001/4/e28/}
\end{botherref}
\endbibitem

\bibitem{Viviani2017}
\begin{barticle}
\bauthor{\bsnm{Viviani}, \binits{M.}},
\bauthor{\bsnm{Pasi}, \binits{G.}}:
\batitle{{Credibility in social media: opinions, news, and health information-a
  survey}}.
\bjtitle{Wiley Interdisciplinary Reviews: Data Mining and Knowledge Discovery}
\bvolume{7}(\bissue{5}),
\bfpage{1209}
(\byear{2017}).
doi:\doiurl{10.1002/widm.1209}
\end{barticle}
\endbibitem

\bibitem{Charnock1999}
\begin{barticle}
\bauthor{\bsnm{Charnock}, \binits{D.}},
\bauthor{\bsnm{Shepperd}, \binits{S.}},
\bauthor{\bsnm{Needham}, \binits{G.}},
\bauthor{\bsnm{Gann}, \binits{R.}}:
\batitle{{DISCERN: an instrument for judging the quality of written consumer
  health information on treatment choices.}}
\bjtitle{Journal of epidemiology and community health}
\bvolume{53}(\bissue{2}),
\bfpage{105}--\blpage{11}
(\byear{1999}).
doi:\doiurl{10.1136/jech.53.2.105}
\end{barticle}
\endbibitem

\bibitem{Boyer2014}
\begin{botherref}
\oauthor{\bsnm{Boyer}, \binits{C.}},
\oauthor{\bsnm{Dolamic}, \binits{L.}}:
{Feasibility of automated detection of HONcode conformity for health-related
  websites}.
International Journal of Advanced Computer Science and Applications
\textbf{5}(3)
(2014).
doi:\doiurl{10.14569/IJACSA.2014.050309}
\end{botherref}
\endbibitem

\bibitem{Boyer2015}
\begin{barticle}
\bauthor{\bsnm{Boyer}, \binits{C.}},
\bauthor{\bsnm{Dolamic}, \binits{L.}}:
\batitle{{Automated Detection of HONcode Website Conformity Compared to Manual
  Detection: An Evaluation.}}
\bjtitle{Journal of medical Internet research}
\bvolume{17}(\bissue{6}),
\bfpage{135}
(\byear{2015}).
doi:\doiurl{10.2196/jmir.3831}
\end{barticle}
\endbibitem

\bibitem{Vaswani2017}
\begin{botherref}
\oauthor{\bsnm{Vaswani}, \binits{A.}},
\oauthor{\bsnm{Shazeer}, \binits{N.}},
\oauthor{\bsnm{Parmar}, \binits{N.}},
\oauthor{\bsnm{Uszkoreit}, \binits{J.}},
\oauthor{\bsnm{Jones}, \binits{L.}},
\oauthor{\bsnm{Gomez}, \binits{A.N.}},
\oauthor{\bsnm{Kaiser}, \binits{L.}},
\oauthor{\bsnm{Polosukhin}, \binits{I.}}:
{Attention Is All You Need}
(2017).
\arxivurl{1706.03762}
\end{botherref}
\endbibitem

\bibitem{Luong2015}
\begin{botherref}
\oauthor{\bsnm{Luong}, \binits{M.-T.}},
\oauthor{\bsnm{Pham}, \binits{H.}},
\oauthor{\bsnm{Manning}, \binits{C.D.}}:
{Effective Approaches to Attention-based Neural Machine Translation}
(2015).
\arxivurl{1508.04025}
\end{botherref}
\endbibitem

\bibitem{Wolf2019}
\begin{botherref}
\oauthor{\bsnm{Wolf}, \binits{T.}},
\oauthor{\bsnm{Debut}, \binits{L.}},
\oauthor{\bsnm{Sanh}, \binits{V.}},
\oauthor{\bsnm{Chaumond}, \binits{J.}},
\oauthor{\bsnm{Delangue}, \binits{C.}},
\oauthor{\bsnm{Moi}, \binits{A.}},
\oauthor{\bsnm{Cistac}, \binits{P.}},
\oauthor{\bsnm{Rault}, \binits{T.}},
\oauthor{\bsnm{Louf}, \binits{R.}},
\oauthor{\bsnm{Funtowicz}, \binits{M.}},
\oauthor{\bsnm{Brew}, \binits{J.}}:
{Transformers: State-of-the-art Natural Language Processing}
(2019).
\arxivurl{1910.03771}
\end{botherref}
\endbibitem

\bibitem{Ruder2019}
\begin{bchapter}
\bauthor{\bsnm{Ruder}, \binits{S.}},
\bauthor{\bsnm{Peters}, \binits{M.E.}},
\bauthor{\bsnm{Swayamdipta}, \binits{S.}},
\bauthor{\bsnm{Wolf}, \binits{T.}}:
\bctitle{{Transfer Learning in Natural Language Processing}}.
In: \bbtitle{Proceedings of the 2019 Conference of the North},
pp. \bfpage{15}--\blpage{18}.
\bpublisher{Association for Computational Linguistics},
\blocation{Stroudsburg, PA, USA}
(\byear{2019}).
doi:\doiurl{10.18653/v1/N19-5004}.
\burl{http://aclweb.org/anthology/N19-5004}
\end{bchapter}
\endbibitem

\bibitem{Rees2002}
\begin{barticle}
\bauthor{\bsnm{Rees}, \binits{C.E.}},
\bauthor{\bsnm{Ford}, \binits{J.E.}},
\bauthor{\bsnm{Sheard}, \binits{C.E.}}:
\batitle{{Evaluating the reliability of DISCERN: A tool for assessing the
  quality of written patient information on treatment choices}}.
\bjtitle{Patient Education and Counseling}
(\byear{2002}).
doi:\doiurl{10.1016/S0738-3991(01)00225-7}
\end{barticle}
\endbibitem

\bibitem{Khazaal2009}
\begin{barticle}
\bauthor{\bsnm{Khazaal}, \binits{Y.}},
\bauthor{\bsnm{Chatton}, \binits{A.}},
\bauthor{\bsnm{Cochand}, \binits{S.}},
\bauthor{\bsnm{Coquard}, \binits{O.}},
\bauthor{\bsnm{Fernandez}, \binits{S.}},
\bauthor{\bsnm{Khan}, \binits{R.}},
\bauthor{\bsnm{Billieux}, \binits{J.}},
\bauthor{\bsnm{Zullino}, \binits{D.}}:
\batitle{{Brief DISCERN, six questions for the evaluation of evidence-based
  content of health-related websites}}.
\bjtitle{Patient Education and Counseling}
(\byear{2009}).
doi:\doiurl{10.1016/j.pec.2009.02.016}
\end{barticle}
\endbibitem

\bibitem{devlin2018bert}
\begin{botherref}
\oauthor{\bsnm{Devlin}, \binits{J.}},
\oauthor{\bsnm{Chang}, \binits{M.-W.}},
\oauthor{\bsnm{Lee}, \binits{K.}},
\oauthor{\bsnm{Toutanova}, \binits{K.}}:
BERT: Pre-training of Deep Bidirectional Transformers for Language
  Understanding
(2018).
\arxivurl{1810.04805}
\end{botherref}
\endbibitem

\bibitem{10.1093/bioinformatics/btz682}
\begin{barticle}
\bauthor{\bsnm{Lee}, \binits{J.}},
\bauthor{\bsnm{Yoon}, \binits{W.}},
\bauthor{\bsnm{Kim}, \binits{S.}},
\bauthor{\bsnm{Kim}, \binits{D.}},
\bauthor{\bsnm{Kim}, \binits{S.}},
\bauthor{\bsnm{So}, \binits{C.H.}},
\bauthor{\bsnm{Kang}, \binits{J.}}:
\batitle{{BioBERT: a pre-trained biomedical language representation model for
  biomedical text mining}}.
\bjtitle{Bioinformatics}
(\byear{2019}).
doi:\doiurl{10.1093/bioinformatics/btz682}
\end{barticle}
\endbibitem

\bibitem{Allam}
\begin{barticle}
\bauthor{\bsnm{Allam}, \binits{A.}},
\bauthor{\bsnm{Schulz}, \binits{P.J.}},
\bauthor{\bsnm{Krauthammer}, \binits{M.}}:
\batitle{{Toward automated assessment of health Web page quality using the
  DISCERN instrument}}.
\bjtitle{Journal of the American Medical Informatics Association}
\bvolume{24}(\bissue{3}),
\bfpage{481}--\blpage{487}
(\byear{2017}).
doi:\doiurl{10.1093/jamia/ocw140}
\end{barticle}
\endbibitem

\bibitem{richardson2007beautiful}
\begin{botherref}
\oauthor{\bsnm{Richardson}, \binits{L.}}:
Beautiful soup documentation.
April
(2007)
\end{botherref}
\endbibitem

\bibitem{Paszke2017AutomaticDI}
\begin{bchapter}
\bauthor{\bsnm{Paszke}, \binits{A.}},
\bauthor{\bsnm{Gross}, \binits{S.}},
\bauthor{\bsnm{Chintala}, \binits{S.}},
\bauthor{\bsnm{Chanan}, \binits{G.}},
\bauthor{\bsnm{Yang}, \binits{E.}},
\bauthor{\bsnm{Devito}, \binits{Z.}},
\bauthor{\bsnm{Lin}, \binits{Z.}},
\bauthor{\bsnm{Desmaison}, \binits{A.}},
\bauthor{\bsnm{Antiga}, \binits{L.}},
\bauthor{\bsnm{Lerer}, \binits{A.}}:
\bctitle{Automatic differentiation in pytorch}.
(\byear{2017})
\end{bchapter}
\endbibitem

\bibitem{Hochreiter1997}
\begin{barticle}
\bauthor{\bsnm{Hochreiter}, \binits{S.}},
\bauthor{\bsnm{Schmidhuber}, \binits{J.}}:
\batitle{{Long Short-Term Memory}}.
\bjtitle{Neural Computation}
\bvolume{9}(\bissue{8}),
\bfpage{1735}--\blpage{1780}
(\byear{1997}).
doi:\doiurl{10.1162/neco.1997.9.8.1735}
\end{barticle}
\endbibitem

\bibitem{Bengio1994}
\begin{barticle}
\bauthor{\bsnm{Bengio}, \binits{Y.}},
\bauthor{\bsnm{Simard}, \binits{P.}},
\bauthor{\bsnm{Frasconi}, \binits{P.}}:
\batitle{{Learning long-term dependencies with gradient descent is difficult}}.
\bjtitle{IEEE Transactions on Neural Networks}
\bvolume{5}(\bissue{2}),
\bfpage{157}--\blpage{166}
(\byear{1994}).
doi:\doiurl{10.1109/72.279181}
\end{barticle}
\endbibitem

\bibitem{Graves2012}
\begin{bbook}
\bauthor{\bsnm{Graves}, \binits{A.}}:
\bbtitle{Supervised Sequence Labelling with Recurrent Neural Networks}.
\bsertitle{Studies in Computational Intelligence},
vol. \bseriesno{385}.
\bpublisher{Springer},
\blocation{Berlin, Heidelberg}
(\byear{2012}).
doi:\doiurl{10.1007/978-3-642-24797-2}.
\burl{http://link.springer.com/10.1007/978-3-642-24797-2}
\end{bbook}
\endbibitem

\bibitem{Cho2014}
\begin{bchapter}
\bauthor{\bsnm{Cho}, \binits{K.}},
\bauthor{\bsnm{{Van Merri{\"{e}}nboer}}, \binits{B.}},
\bauthor{\bsnm{Gulcehre}, \binits{C.}},
\bauthor{\bsnm{Bahdanau}, \binits{D.}},
\bauthor{\bsnm{Bougares}, \binits{F.}},
\bauthor{\bsnm{Schwenk}, \binits{H.}},
\bauthor{\bsnm{Bengio}, \binits{Y.}}:
\bctitle{{Learning Phrase Representations using RNN Encoder-Decoder for
  Statistical Machine Translation}}.
In: \bbtitle{Proceedings of the 2014 Conference on Empirical Methods in Natural
  Language Processing (EMNLP)},
pp. \bfpage{1724}--\blpage{1734}.
\bpublisher{Association for Computational Linguistics},
\blocation{Doha, Qatar}
(\byear{2014}).
\burl{http://aclweb.org/anthology/D14-1179}
\end{bchapter}
\endbibitem

\bibitem{chung2014empirical}
\begin{botherref}
\oauthor{\bsnm{Chung}, \binits{J.}},
\oauthor{\bsnm{Gulcehre}, \binits{C.}},
\oauthor{\bsnm{Cho}, \binits{K.}},
\oauthor{\bsnm{Bengio}, \binits{Y.}}:
{Empirical Evaluation of Gated Recurrent Neural Networks on Sequence Modeling}
(2014).
\arxivurl{1412.3555}
\end{botherref}
\endbibitem

\bibitem{Bahdanau2014a}
\begin{botherref}
\oauthor{\bsnm{Bahdanau}, \binits{D.}},
\oauthor{\bsnm{Cho}, \binits{K.}},
\oauthor{\bsnm{Bengio}, \binits{Y.}}:
{Neural Machine Translation by Jointly Learning to Align and Translate}
(2014).
\arxivurl{1409.0473}
\end{botherref}
\endbibitem

\bibitem{Srivastava2014}
\begin{barticle}
\bauthor{\bsnm{Srivastava}, \binits{N.}},
\bauthor{\bsnm{Hinton}, \binits{G.}},
\bauthor{\bsnm{Krizhevsky}, \binits{A.}},
\bauthor{\bsnm{Sutskever}, \binits{I.}},
\bauthor{\bsnm{Salakhutdinov}, \binits{R.}}:
\batitle{{Dropout: A Simple Way to Prevent Neural Networks from Overfitting}}.
\bjtitle{Journal of Machine Learning Research}
\bvolume{15},
\bfpage{1929}--\blpage{1958}
(\byear{2014})
\end{barticle}
\endbibitem

\bibitem{BergstraJAMESBERGSTRA2012}
\begin{barticle}
\bauthor{\bsnm{{Bergstra JAMESBERGSTRA}}, \binits{J.}},
\bauthor{\bsnm{{Yoshua Bengio YOSHUABENGIO}}, \binits{U.}}:
\batitle{{Random Search for HyperParameter Optimization}}.
\bjtitle{Journal of Machine Learning Research}
(\byear{2012}).
doi:\doiurl{10.1162/153244303322533223}.
\arxivurl{1504.05070}
\end{barticle}
\endbibitem

\bibitem{Demner-Fushman2017}
\begin{botherref}
\oauthor{\bsnm{Demner-Fushman}, \binits{D.}},
\oauthor{\bsnm{Rogers}, \binits{W.J.}},
\oauthor{\bsnm{Aronson}, \binits{A.R.}}:
{MetaMap Lite: an evaluation of a new Java implementation of MetaMap}.
Journal of the American Medical Informatics Association,
177
(2017).
doi:\doiurl{10.1093/jamia/ocw177}
\end{botherref}
\endbibitem

\bibitem{scikit-learn}
\begin{barticle}
\bauthor{\bsnm{Pedregosa}, \binits{F.}},
\bauthor{\bsnm{Varoquaux}, \binits{G.}},
\bauthor{\bsnm{Gramfort}, \binits{A.}},
\bauthor{\bsnm{Michel}, \binits{V.}},
\bauthor{\bsnm{Thirion}, \binits{B.}},
\bauthor{\bsnm{Grisel}, \binits{O.}},
\bauthor{\bsnm{Blondel}, \binits{M.}},
\bauthor{\bsnm{Prettenhofer}, \binits{P.}},
\bauthor{\bsnm{Weiss}, \binits{R.}},
\bauthor{\bsnm{Dubourg}, \binits{V.}},
\bauthor{\bsnm{Vanderplas}, \binits{J.}},
\bauthor{\bsnm{Passos}, \binits{A.}},
\bauthor{\bsnm{Cournapeau}, \binits{D.}},
\bauthor{\bsnm{Brucher}, \binits{M.}},
\bauthor{\bsnm{Perrot}, \binits{M.}},
\bauthor{\bsnm{Duchesnay}, \binits{E.}}:
\batitle{Scikit-learn: Machine learning in {P}ython}.
\bjtitle{Journal of Machine Learning Research}
\bvolume{12},
\bfpage{2825}--\blpage{2830}
(\byear{2011})
\end{barticle}
\endbibitem

\bibitem{Berthelot2019}
\begin{botherref}
\oauthor{\bsnm{Berthelot}, \binits{D.}},
\oauthor{\bsnm{Carlini}, \binits{N.}},
\oauthor{\bsnm{Goodfellow}, \binits{I.}},
\oauthor{\bsnm{Papernot}, \binits{N.}},
\oauthor{\bsnm{Oliver}, \binits{A.}},
\oauthor{\bsnm{Raffel}, \binits{C.}}:
{MixMatch: A Holistic Approach to Semi-Supervised Learning}
(2019).
\arxivurl{1905.02249}
\end{botherref}
\endbibitem

\bibitem{Xie2019}
\begin{botherref}
\oauthor{\bsnm{Xie}, \binits{Q.}},
\oauthor{\bsnm{Dai}, \binits{Z.}},
\oauthor{\bsnm{Hovy}, \binits{E.}},
\oauthor{\bsnm{Luong}, \binits{M.-T.}},
\oauthor{\bsnm{Le}, \binits{Q.V.}}:
{Unsupervised Data Augmentation for Consistency Training}
(2019).
\arxivurl{1904.12848}
\end{botherref}
\endbibitem

\end{thebibliography}

\newcommand{\BMCxmlcomment}[1]{}

\BMCxmlcomment{

<refgrp>

<bibl id="B1">
  <title><p>{Trust and Sources of Health Information}</p></title>
  <aug>
    <au><snm>Hesse</snm><fnm>BW</fnm></au>
    <au><snm>Nelson</snm><fnm>DE</fnm></au>
    <au><snm>Kreps</snm><fnm>GL</fnm></au>
    <au><snm>Croyle</snm><fnm>RT</fnm></au>
    <au><snm>Arora</snm><fnm>NK</fnm></au>
    <au><snm>Rimer</snm><fnm>BK</fnm></au>
    <au><snm>Viswanath</snm><fnm>K</fnm></au>
  </aug>
  <source>Archives of Internal Medicine</source>
  <pubdate>2005</pubdate>
  <volume>165</volume>
  <issue>22</issue>
  <fpage>2618</fpage>
  <url>http://www.ncbi.nlm.nih.gov/pubmed/16344419</url>
</bibl>

<bibl id="B2">
  <title><p>{Quality of patient health information on the internet: reviewing a
  complex and}</p></title>
  <aug>
    <au><snm>Fahy</snm><fnm>E</fnm></au>
  </aug>
  <source>Australasian Medical Journal</source>
  <pubdate>2014</pubdate>
  <volume>7</volume>
  <issue>1</issue>
  <fpage>24</fpage>
  <lpage>-28</lpage>
</bibl>

<bibl id="B3">
  <title><p>{Quality of health information for consumers on the web: A
  systematic review of indicators, criteria, tools, and evaluation
  results}</p></title>
  <aug>
    <au><snm>Zhang</snm><fnm>Y</fnm></au>
    <au><snm>Sun</snm><fnm>Y</fnm></au>
    <au><snm>Xie</snm><fnm>B</fnm></au>
  </aug>
  <source>Journal of the Association for Information Science and
  Technology</source>
  <pubdate>2015</pubdate>
  <volume>66</volume>
  <issue>10</issue>
</bibl>

<bibl id="B4">
  <title><p>{Bring on the Machines: Could Machine Learning Improve the Quality
  of Patient Education Materials? A Systematic Search and Rapid
  Review}</p></title>
  <aug>
    <au><snm>Saunders</snm><fnm>CH</fnm></au>
    <au><snm>Petersen</snm><fnm>CL</fnm></au>
    <au><snm>Durang</snm><fnm>MA</fnm></au>
    <au><snm>Bagley</snm><fnm>PJ</fnm></au>
    <au><snm>Elywn</snm><fnm>G</fnm></au>
  </aug>
  <source>American Society of Clinical Oncology</source>
  <pubdate>2018</pubdate>
  <url>https://ascopubs.org/doi/pdfdirect/10.1200/CCI.18.00010</url>
</bibl>

<bibl id="B5">
  <title><p>{The Impact of Health Information on the Internet on Health Care
  and the Physician-Patient Relationship: National U.S. Survey among 1.050 U.S.
  Physicians}</p></title>
  <aug>
    <au><snm>Murray</snm><fnm>E</fnm></au>
    <au><snm>Lo</snm><fnm>B</fnm></au>
    <au><snm>Pollack</snm><fnm>L</fnm></au>
    <au><snm>Donelan</snm><fnm>K</fnm></au>
    <au><snm>Catania</snm><fnm>J</fnm></au>
    <au><snm>Lee</snm><fnm>K</fnm></au>
    <au><snm>Zapert</snm><fnm>K</fnm></au>
    <au><snm>Turner</snm><fnm>R</fnm></au>
  </aug>
  <source>Journal of Medical Internet Research</source>
  <pubdate>2003</pubdate>
  <volume>5</volume>
  <issue>3</issue>
  <fpage>e17</fpage>
  <url>http://www.jmir.org/2003/3/e17/</url>
</bibl>

<bibl id="B6">
  <title><p>{The impact of search engine selection and sorting criteria on
  vaccination beliefs and attitudes: two experiments manipulating Google
  output.}</p></title>
  <aug>
    <au><snm>Allam</snm><fnm>A</fnm></au>
    <au><snm>Schulz</snm><fnm>PJ</fnm></au>
    <au><snm>Nakamoto</snm><fnm>K</fnm></au>
  </aug>
  <source>Journal of medical Internet research</source>
  <publisher>Journal of Medical Internet Research</publisher>
  <pubdate>2014</pubdate>
  <volume>16</volume>
  <issue>4</issue>
  <fpage>e100</fpage>
  <url>http://www.jmir.org/2014/4/e100/</url>
</bibl>

<bibl id="B7">
  <title><p>{Manipulating Google's Knowledge Graph Box to Counter Biased
  Information Processing During an Online Search on Vaccination: Application of
  a Technological Debiasing Strategy}</p></title>
  <aug>
    <au><snm>Ludolph</snm><fnm>R</fnm></au>
    <au><snm>Allam</snm><fnm>A</fnm></au>
    <au><snm>Schulz</snm><fnm>PJ</fnm></au>
  </aug>
  <source>Journal of Medical Internet Research</source>
  <publisher>Journal of Medical Internet Research</publisher>
  <pubdate>2016</pubdate>
  <volume>18</volume>
  <issue>6</issue>
  <fpage>e137</fpage>
  <url>http://www.jmir.org/2016/6/e137/</url>
</bibl>

<bibl id="B8">
  <title><p>Impact of internet use on health-related behaviors and the
  patient-physician relationship: a survey-based study and review.</p></title>
  <aug>
    <au><snm>Iverson</snm><fnm>SA</fnm></au>
    <au><snm>Howard</snm><fnm>KB</fnm></au>
    <au><snm>Penney</snm><fnm>BK</fnm></au>
  </aug>
  <source>The Journal of the American Osteopathic Association</source>
  <pubdate>2008</pubdate>
  <volume>108</volume>
  <issue>12</issue>
  <fpage>699</fpage>
  <lpage>-711</lpage>
  <url>http://www.ncbi.nlm.nih.gov/pubmed/19075034</url>
</bibl>

<bibl id="B9">
  <title><p>{Untangling the Web-The impact of Internet use on health care and
  the physician-patient relationship}</p></title>
  <aug>
    <au><snm>Wald</snm><fnm>HS</fnm></au>
    <au><snm>Dube</snm><fnm>CE</fnm></au>
    <au><snm>Anthony</snm><fnm>DC</fnm></au>
  </aug>
  <source>Patient Education and Counseling</source>
  <pubdate>2007</pubdate>
  <volume>68</volume>
  <issue>3</issue>
  <fpage>218</fpage>
  <lpage>-224</lpage>
</bibl>

<bibl id="B10">
  <title><p>{Review of Internet health information quality
  initiatives}</p></title>
  <aug>
    <au><snm>Risk</snm><fnm>A</fnm></au>
    <au><snm>Dzenowagis</snm><fnm>J</fnm></au>
  </aug>
  <source>Journal of Medical Internet Research</source>
  <publisher>Journal of Medical Internet Research</publisher>
  <pubdate>2001</pubdate>
  <volume>3</volume>
  <issue>4</issue>
  <fpage>17</fpage>
  <lpage>-45</lpage>
  <url>http://www.jmir.org/2001/4/e28/</url>
</bibl>

<bibl id="B11">
  <title><p>{Credibility in social media: opinions, news, and health
  information-a survey}</p></title>
  <aug>
    <au><snm>Viviani</snm><fnm>M</fnm></au>
    <au><snm>Pasi</snm><fnm>G</fnm></au>
  </aug>
  <source>Wiley Interdisciplinary Reviews: Data Mining and Knowledge
  Discovery</source>
  <publisher>Wiley-Blackwell</publisher>
  <pubdate>2017</pubdate>
  <volume>7</volume>
  <issue>5</issue>
  <fpage>e1209</fpage>
  <url>http://doi.wiley.com/10.1002/widm.1209</url>
</bibl>

<bibl id="B12">
  <title><p>{DISCERN: an instrument for judging the quality of written consumer
  health information on treatment choices.}</p></title>
  <aug>
    <au><snm>Charnock</snm><fnm>D</fnm></au>
    <au><snm>Shepperd</snm><fnm>S</fnm></au>
    <au><snm>Needham</snm><fnm>G</fnm></au>
    <au><snm>Gann</snm><fnm>R</fnm></au>
  </aug>
  <source>Journal of epidemiology and community health</source>
  <publisher>BMJ Publishing Group</publisher>
  <pubdate>1999</pubdate>
  <volume>53</volume>
  <issue>2</issue>
  <fpage>105</fpage>
  <lpage>-11</lpage>
  <url>http://www.ncbi.nlm.nih.gov/pubmed/10396471</url>
</bibl>

<bibl id="B13">
  <title><p>{Feasibility of automated detection of HONcode conformity for
  health-related websites}</p></title>
  <aug>
    <au><snm>Boyer</snm><fnm>C</fnm></au>
    <au><snm>Dolamic</snm><fnm>L</fnm></au>
  </aug>
  <source>International Journal of Advanced Computer Science and
  Applications</source>
  <pubdate>2014</pubdate>
  <volume>5</volume>
  <issue>3</issue>
  <url>http://thesai.org/Publications/ViewPaper?Volume=5{\&}Issue=3{\&}Code=IJACSA{\&}SerialNo=9</url>
</bibl>

<bibl id="B14">
  <title><p>{Automated Detection of HONcode Website Conformity Compared to
  Manual Detection: An Evaluation.}</p></title>
  <aug>
    <au><snm>Boyer</snm><fnm>C</fnm></au>
    <au><snm>Dolamic</snm><fnm>L</fnm></au>
  </aug>
  <source>Journal of medical Internet research</source>
  <publisher>JMIR Publications Inc.</publisher>
  <pubdate>2015</pubdate>
  <volume>17</volume>
  <issue>6</issue>
  <fpage>e135</fpage>
  <url>http://www.ncbi.nlm.nih.gov/pubmed/26036669</url>
</bibl>

<bibl id="B15">
  <title><p>{Attention Is All You Need}</p></title>
  <aug>
    <au><snm>Vaswani</snm><fnm>A</fnm></au>
    <au><snm>Shazeer</snm><fnm>N</fnm></au>
    <au><snm>Parmar</snm><fnm>N</fnm></au>
    <au><snm>Uszkoreit</snm><fnm>J</fnm></au>
    <au><snm>Jones</snm><fnm>L</fnm></au>
    <au><snm>Gomez</snm><fnm>AN</fnm></au>
    <au><snm>Kaiser</snm><fnm>L</fnm></au>
    <au><snm>Polosukhin</snm><fnm>I</fnm></au>
  </aug>
  <pubdate>2017</pubdate>
  <url>http://arxiv.org/abs/1706.03762</url>
</bibl>

<bibl id="B16">
  <title><p>{Effective Approaches to Attention-based Neural Machine
  Translation}</p></title>
  <aug>
    <au><snm>Luong</snm><fnm>MT</fnm></au>
    <au><snm>Pham</snm><fnm>H</fnm></au>
    <au><snm>Manning</snm><fnm>CD</fnm></au>
  </aug>
  <pubdate>2015</pubdate>
  <url>http://arxiv.org/abs/1508.04025</url>
</bibl>

<bibl id="B17">
  <title><p>{Transformers: State-of-the-art Natural Language
  Processing}</p></title>
  <aug>
    <au><snm>Wolf</snm><fnm>T</fnm></au>
    <au><snm>Debut</snm><fnm>L</fnm></au>
    <au><snm>Sanh</snm><fnm>V</fnm></au>
    <au><snm>Chaumond</snm><fnm>J</fnm></au>
    <au><snm>Delangue</snm><fnm>C</fnm></au>
    <au><snm>Moi</snm><fnm>A</fnm></au>
    <au><snm>Cistac</snm><fnm>P</fnm></au>
    <au><snm>Rault</snm><fnm>T</fnm></au>
    <au><snm>Louf</snm><fnm>R</fnm></au>
    <au><snm>Funtowicz</snm><fnm>M</fnm></au>
    <au><snm>Brew</snm><fnm>J</fnm></au>
  </aug>
  <pubdate>2019</pubdate>
  <url>http://arxiv.org/abs/1910.03771</url>
</bibl>

<bibl id="B18">
  <title><p>{Transfer Learning in Natural Language Processing}</p></title>
  <aug>
    <au><snm>Ruder</snm><fnm>S</fnm></au>
    <au><snm>Peters</snm><fnm>ME</fnm></au>
    <au><snm>Swayamdipta</snm><fnm>S</fnm></au>
    <au><snm>Wolf</snm><fnm>T</fnm></au>
  </aug>
  <source>Proceedings of the 2019 Conference of the North</source>
  <publisher>Stroudsburg, PA, USA: Association for Computational
  Linguistics</publisher>
  <pubdate>2019</pubdate>
  <fpage>15</fpage>
  <lpage>-18</lpage>
  <url>http://aclweb.org/anthology/N19-5004</url>
</bibl>

<bibl id="B19">
  <title><p>{Evaluating the reliability of DISCERN: A tool for assessing the
  quality of written patient information on treatment choices}</p></title>
  <aug>
    <au><snm>Rees</snm><fnm>CE</fnm></au>
    <au><snm>Ford</snm><fnm>JE</fnm></au>
    <au><snm>Sheard</snm><fnm>CE</fnm></au>
  </aug>
  <source>Patient Education and Counseling</source>
  <pubdate>2002</pubdate>
</bibl>

<bibl id="B20">
  <title><p>{Brief DISCERN, six questions for the evaluation of evidence-based
  content of health-related websites}</p></title>
  <aug>
    <au><snm>Khazaal</snm><fnm>Y</fnm></au>
    <au><snm>Chatton</snm><fnm>A</fnm></au>
    <au><snm>Cochand</snm><fnm>S</fnm></au>
    <au><snm>Coquard</snm><fnm>O</fnm></au>
    <au><snm>Fernandez</snm><fnm>S</fnm></au>
    <au><snm>Khan</snm><fnm>R</fnm></au>
    <au><snm>Billieux</snm><fnm>J</fnm></au>
    <au><snm>Zullino</snm><fnm>D</fnm></au>
  </aug>
  <source>Patient Education and Counseling</source>
  <pubdate>2009</pubdate>
</bibl>

<bibl id="B21">
  <title><p>BERT: Pre-training of Deep Bidirectional Transformers for Language
  Understanding</p></title>
  <aug>
    <au><snm>Devlin</snm><fnm>J</fnm></au>
    <au><snm>Chang</snm><fnm>MW</fnm></au>
    <au><snm>Lee</snm><fnm>K</fnm></au>
    <au><snm>Toutanova</snm><fnm>K</fnm></au>
  </aug>
  <pubdate>2018</pubdate>
</bibl>

<bibl id="B22">
  <title><p>{BioBERT: a pre-trained biomedical language representation model
  for biomedical text mining}</p></title>
  <aug>
    <au><snm>Lee</snm><fnm>J</fnm></au>
    <au><snm>Yoon</snm><fnm>W</fnm></au>
    <au><snm>Kim</snm><fnm>S</fnm></au>
    <au><snm>Kim</snm><fnm>D</fnm></au>
    <au><snm>Kim</snm><fnm>S</fnm></au>
    <au><snm>So</snm><fnm>CH</fnm></au>
    <au><snm>Kang</snm><fnm>J</fnm></au>
  </aug>
  <source>Bioinformatics</source>
  <pubdate>2019</pubdate>
  <url>https://doi.org/10.1093/bioinformatics/btz682</url>
</bibl>

<bibl id="B23">
  <title><p>{Toward automated assessment of health Web page quality using the
  DISCERN instrument}</p></title>
  <aug>
    <au><snm>Allam</snm><fnm>A</fnm></au>
    <au><snm>Schulz</snm><fnm>PJ</fnm></au>
    <au><snm>Krauthammer</snm><fnm>M</fnm></au>
  </aug>
  <source>Journal of the American Medical Informatics Association</source>
  <pubdate>2017</pubdate>
  <volume>24</volume>
  <issue>3</issue>
  <fpage>481</fpage>
  <lpage>-487</lpage>
</bibl>

<bibl id="B24">
  <title><p>Beautiful soup documentation</p></title>
  <aug>
    <au><snm>Richardson</snm><fnm>L</fnm></au>
  </aug>
  <source>April</source>
  <pubdate>2007</pubdate>
</bibl>

<bibl id="B25">
  <title><p>Automatic differentiation in PyTorch</p></title>
  <aug>
    <au><snm>Paszke</snm><fnm>A</fnm></au>
    <au><snm>Gross</snm><fnm>S</fnm></au>
    <au><snm>Chintala</snm><fnm>S</fnm></au>
    <au><snm>Chanan</snm><fnm>G</fnm></au>
    <au><snm>Yang</snm><fnm>E</fnm></au>
    <au><snm>Devito</snm><fnm>Z</fnm></au>
    <au><snm>Lin</snm><fnm>Z</fnm></au>
    <au><snm>Desmaison</snm><fnm>A</fnm></au>
    <au><snm>Antiga</snm><fnm>L</fnm></au>
    <au><snm>Lerer</snm><fnm>A</fnm></au>
  </aug>
  <pubdate>2017</pubdate>
</bibl>

<bibl id="B26">
  <title><p>{Long Short-Term Memory}</p></title>
  <aug>
    <au><snm>Hochreiter</snm><fnm>S</fnm></au>
    <au><snm>Schmidhuber</snm><fnm>J</fnm></au>
  </aug>
  <source>Neural Computation</source>
  <publisher>MIT Press 238 Main St., Suite 500, Cambridge, MA 02142-1046 USA
  journals-info@mit.edu</publisher>
  <pubdate>1997</pubdate>
  <volume>9</volume>
  <issue>8</issue>
  <fpage>1735</fpage>
  <lpage>-1780</lpage>
  <url>http://www.mitpressjournals.org/doi/10.1162/neco.1997.9.8.1735</url>
</bibl>

<bibl id="B27">
  <title><p>{Learning long-term dependencies with gradient descent is
  difficult}</p></title>
  <aug>
    <au><snm>Bengio</snm><fnm>Y.</fnm></au>
    <au><snm>Simard</snm><fnm>P.</fnm></au>
    <au><snm>Frasconi</snm><fnm>P.</fnm></au>
  </aug>
  <source>IEEE Transactions on Neural Networks</source>
  <pubdate>1994</pubdate>
  <volume>5</volume>
  <issue>2</issue>
  <fpage>157</fpage>
  <lpage>-166</lpage>
  <url>http://ieeexplore.ieee.org/document/279181/</url>
</bibl>

<bibl id="B28">
  <title><p>Supervised Sequence Labelling with Recurrent Neural
  Networks</p></title>
  <aug>
    <au><snm>Graves</snm><fnm>A</fnm></au>
  </aug>
  <publisher>Berlin, Heidelberg: Springer Berlin Heidelberg</publisher>
  <series><title><p>Studies in Computational Intelligence</p></title></series>
  <pubdate>2012</pubdate>
  <volume>385</volume>
  <url>http://link.springer.com/10.1007/978-3-642-24797-2</url>
</bibl>

<bibl id="B29">
  <title><p>{Learning Phrase Representations using RNN Encoder-Decoder for
  Statistical Machine Translation}</p></title>
  <aug>
    <au><snm>Cho</snm><fnm>K</fnm></au>
    <au><snm>{Van Merri{\"{e}}nboer}</snm><fnm>B</fnm></au>
    <au><snm>Gulcehre</snm><fnm>C</fnm></au>
    <au><snm>Bahdanau</snm><fnm>D</fnm></au>
    <au><snm>Bougares</snm><fnm>F</fnm></au>
    <au><snm>Schwenk</snm><fnm>H</fnm></au>
    <au><snm>Bengio</snm><fnm>Y</fnm></au>
  </aug>
  <source>Proceedings of the 2014 Conference on Empirical Methods in Natural
  Language Processing (EMNLP)</source>
  <publisher>Doha, Qatar: Association for Computational Linguistics</publisher>
  <pubdate>2014</pubdate>
  <fpage>1724</fpage>
  <lpage>-1734</lpage>
  <url>http://aclweb.org/anthology/D14-1179</url>
</bibl>

<bibl id="B30">
  <title><p>{Empirical Evaluation of Gated Recurrent Neural Networks on
  Sequence Modeling}</p></title>
  <aug>
    <au><snm>Chung</snm><fnm>J</fnm></au>
    <au><snm>Gulcehre</snm><fnm>C</fnm></au>
    <au><snm>Cho</snm><fnm>K</fnm></au>
    <au><snm>Bengio</snm><fnm>Y</fnm></au>
  </aug>
  <pubdate>2014</pubdate>
</bibl>

<bibl id="B31">
  <title><p>{Neural Machine Translation by Jointly Learning to Align and
  Translate}</p></title>
  <aug>
    <au><snm>Bahdanau</snm><fnm>D</fnm></au>
    <au><snm>Cho</snm><fnm>K</fnm></au>
    <au><snm>Bengio</snm><fnm>Y</fnm></au>
  </aug>
  <pubdate>2014</pubdate>
  <url>http://arxiv.org/abs/1409.0473</url>
</bibl>

<bibl id="B32">
  <title><p>{Dropout: A Simple Way to Prevent Neural Networks from
  Overfitting}</p></title>
  <aug>
    <au><snm>Srivastava</snm><fnm>N</fnm></au>
    <au><snm>Hinton</snm><fnm>G</fnm></au>
    <au><snm>Krizhevsky</snm><fnm>A</fnm></au>
    <au><snm>Sutskever</snm><fnm>I</fnm></au>
    <au><snm>Salakhutdinov</snm><fnm>R</fnm></au>
  </aug>
  <source>Journal of Machine Learning Research</source>
  <pubdate>2014</pubdate>
  <volume>15</volume>
  <fpage>1929</fpage>
  <lpage>-1958</lpage>
  <url>http://jmlr.org/papers/v15/srivastava14a.html</url>
</bibl>

<bibl id="B33">
  <title><p>{Random Search for HyperParameter Optimization}</p></title>
  <aug>
    <au><snm>{Bergstra JAMESBERGSTRA}</snm><fnm>J</fnm></au>
    <au><snm>{Yoshua Bengio YOSHUABENGIO}</snm><fnm>U</fnm></au>
  </aug>
  <source>Journal of Machine Learning Research</source>
  <pubdate>2012</pubdate>
</bibl>

<bibl id="B34">
  <title><p>{MetaMap Lite: an evaluation of a new Java implementation of
  MetaMap}</p></title>
  <aug>
    <au><snm>Demner Fushman</snm><fnm>D</fnm></au>
    <au><snm>Rogers</snm><fnm>WJ</fnm></au>
    <au><snm>Aronson</snm><fnm>AR</fnm></au>
  </aug>
  <source>Journal of the American Medical Informatics Association</source>
  <pubdate>2017</pubdate>
  <fpage>ocw177</fpage>
  <url>https://academic.oup.com/jamia/article-lookup/doi/10.1093/jamia/ocw177</url>
</bibl>

<bibl id="B35">
  <title><p>Scikit-learn: Machine Learning in {P}ython</p></title>
  <aug>
    <au><snm>Pedregosa</snm><fnm>F.</fnm></au>
    <au><snm>Varoquaux</snm><fnm>G.</fnm></au>
    <au><snm>Gramfort</snm><fnm>A.</fnm></au>
    <au><snm>Michel</snm><fnm>V.</fnm></au>
    <au><snm>Thirion</snm><fnm>B.</fnm></au>
    <au><snm>Grisel</snm><fnm>O.</fnm></au>
    <au><snm>Blondel</snm><fnm>M.</fnm></au>
    <au><snm>Prettenhofer</snm><fnm>P.</fnm></au>
    <au><snm>Weiss</snm><fnm>R.</fnm></au>
    <au><snm>Dubourg</snm><fnm>V.</fnm></au>
    <au><snm>Vanderplas</snm><fnm>J.</fnm></au>
    <au><snm>Passos</snm><fnm>A.</fnm></au>
    <au><snm>Cournapeau</snm><fnm>D.</fnm></au>
    <au><snm>Brucher</snm><fnm>M.</fnm></au>
    <au><snm>Perrot</snm><fnm>M.</fnm></au>
    <au><snm>Duchesnay</snm><fnm>E.</fnm></au>
  </aug>
  <source>Journal of Machine Learning Research</source>
  <pubdate>2011</pubdate>
  <volume>12</volume>
  <fpage>2825</fpage>
  <lpage>-2830</lpage>
</bibl>

<bibl id="B36">
  <title><p>{MixMatch: A Holistic Approach to Semi-Supervised
  Learning}</p></title>
  <aug>
    <au><snm>Berthelot</snm><fnm>D</fnm></au>
    <au><snm>Carlini</snm><fnm>N</fnm></au>
    <au><snm>Goodfellow</snm><fnm>I</fnm></au>
    <au><snm>Papernot</snm><fnm>N</fnm></au>
    <au><snm>Oliver</snm><fnm>A</fnm></au>
    <au><snm>Raffel</snm><fnm>C</fnm></au>
  </aug>
  <pubdate>2019</pubdate>
  <url>http://arxiv.org/abs/1905.02249</url>
</bibl>

<bibl id="B37">
  <title><p>{Unsupervised Data Augmentation for Consistency
  Training}</p></title>
  <aug>
    <au><snm>Xie</snm><fnm>Q</fnm></au>
    <au><snm>Dai</snm><fnm>Z</fnm></au>
    <au><snm>Hovy</snm><fnm>E</fnm></au>
    <au><snm>Luong</snm><fnm>MT</fnm></au>
    <au><snm>Le</snm><fnm>QV</fnm></au>
  </aug>
  <pubdate>2019</pubdate>
  <url>http://arxiv.org/abs/1904.12848</url>
</bibl>

</refgrp>
} 

\end{document}